\documentclass[final,5p,times,twocolumn]{elsarticle}

\graphicspath{ {figures/} }
\usepackage{amssymb}
\usepackage{amsmath}
\usepackage{array,booktabs, makecell, multirow, threeparttable} 
\usepackage{algorithm, algpseudocode} 
\usepackage{subcaption}  

\journal{Knowledge-Based Systems}

\newtheorem{example}{Example}
\begin{document}
	\title{Dependency-based Anomaly Detection: a General Framework and Comprehensive Evaluation}
	
	\begin{frontmatter}
		
		\author[unisa]{Sha Lu\corref{cor1}\fnref{fn1}}
		\ead{Sha.Lu@unisa.edu.au}
		
		\author[unisa]{Lin Liu\corref{cor1}}
		\ead{Lin.Liu@unisa.edu.au}
		
		\author[hfut]{Kui Yu}
		\ead{yukui@hfut.edu.cn}
		
		\author[unisa]{Thuc Duy Le}
		\ead{Thuc.Le@unisa.edu.au}
		
		\author[unisa]{Jixue Liu}
		\ead{Jixue.Liu@unisa.edu.au}
		
		\author[unisa]{Jiuyong Li\corref{cor1}}
		\ead{Jiuyong.Li@unisa.edu.au}
		
		\affiliation[unisa]{organization={University of South Australia},
			addressline={},
			city={Adelaide},
			postcode={5000},
			state={South Australia},
			country={Australia}}
		
		\affiliation[hfut]{organization={Hefei University of Technology},
			addressline={},
			city={Hefei},
			postcode={230000},
			state={Anhui},
			country={China}}
		
		\cortext[cor1]{Corresponding author}
		\fntext[fn1]{First author}
		
		\begin{abstract}
			Anomaly detection is crucial for understanding unusual behaviors in data, as anomalies offer valuable insights. This paper introduces Dependency-based Anomaly Detection (DepAD), a general framework that utilizes variable dependencies to uncover meaningful anomalies with better interpretability. DepAD reframes unsupervised anomaly detection as supervised feature selection and prediction tasks, which allows users to tailor anomaly detection algorithms to their specific problems and data. We extensively evaluate representative off-the-shelf techniques for the DepAD framework. Two DepAD algorithms emerge as all-rounders and superior performers in handling a wide range of datasets compared to nine state-of-the-art anomaly detection methods. Additionally, we demonstrate that DepAD algorithms provide new and insightful interpretations for detected anomalies.
			
		\end{abstract}

		
	\end{frontmatter}
	
	
	\section{Introduction} \label{sec:introduction}
	
	Anomalies are patterns in data that do not conform to a pre-defined notion of normal behavior. They often contain insights about the unusual behaviors or abnormal characteristics of the data generation process, which may imply flaws or misuse of a system. 
	
	The mainstream anomaly detection methods are based on proximity, including distance-based and density-based methods~\cite{thudumu2020comprehensive, aggarwal2016outlier,chandola2009anomaly}. They assume that normal objects are in a dense neighborhood, while anomalies stay far away from other objects or in a sparse neighborhood.
	
	Another line of research in anomaly detection exploits the dependency among variables, assuming normal objects follow the dependency while anomalies do not. Dependency-based methods~\cite{lu2019LoPAD,paulheim2015decomposition} evaluate the anomalousness of objects through how much they deviate from normal dependency possessed by the majority of objects.
	
	Dependency-based approach is fundamentally different from proximity-based approach because it considers the relationship among variables, while proximity-based approach examines the relationship among objects. We use an example to explain the difference between the two approaches.
	
	\begin{example} \textbf{(Dependency-based and proximity-based anomaly detection.)} 
		\label{emp:obesity} 
		The black dots in Figure \ref{fig:obesity} show the height and weight of all the 452 objects (people) taken from the Arrhythmia dataset in the UCI repository~\cite{UciData}. Two objects, $a_1$ with height 162cm and weight 100kg; and $a_2$ with height 200cm and weight 100kg, are added by us and shown as red crosses in the figure. Suppose that we want to identify people with obesity from this dataset.
		
		\begin{figure}[ht]
			\centering
			\includegraphics[width=0.8\columnwidth]{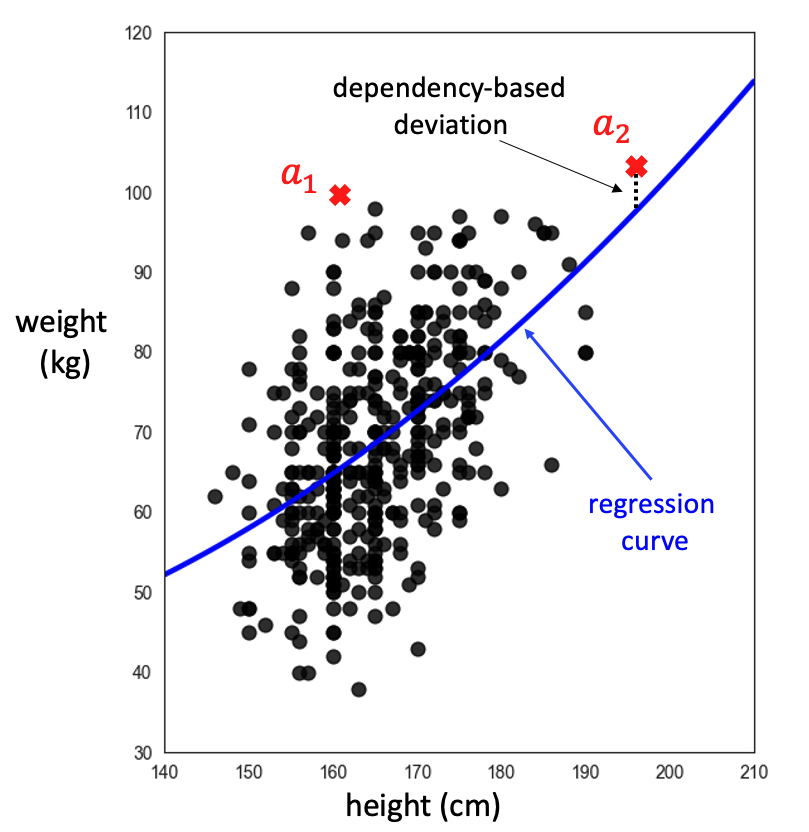} 
			\caption{An example illustrating obesity detection in a dataset with height and weight variables. A proximity-based approach mislabels $a_2$ as obese, while a dependency-based approach correctly identifies $a_1$ as obese and $a_2$ as normal.}
			\label{fig:obesity}
		\end{figure}
		
		As per the World Health Organization's definition of obesity, an adult is considered obese if their body mass index (BMI) exceeds 30, where BMI is calculated as $BMI = weight / height^2$. In this scenario, object $a_1$ is classified as obese, with a BMI of 38, while object $a_2$ is not obese, with a BMI of 25.
		
		When applying a proximity-based method to this dataset, it may incorrectly label object $a_2$ as an anomaly. Proximity-based methods tend to report objects that lie in sparsely populated regions as anomalies, which is why $a_2$ is wrongly identified as such. On the other hand, a dependency-based method analyzes the relationship between height and weight (depicted by the blue curve) and identifies $a_1$ as an anomaly due to its significant deviation from the normal dependency represented by the blue curve. However, $a_2$ is not flagged as an anomaly because its deviation from the blue curve is minimal.
		
		This example highlights the fundamental difference between proximity-based and dependency-based methods. Dependency-based methods focus on identifying anomalies based on underlying relationships between variables, whereas proximity-based methods rely on object similarity in terms of proximity. In cases like this, where the nature of an anomaly is related to the dependency between variables (weight and height), dependency-based methods are more likely to discover meaningful anomalies that proximity-based methods may overlook.
	\end{example}
	
	A common way of examining dependency deviations in the dependency-based approach is to check the difference between the observed value and the expected value of an object, where the expected value is estimated based on the underlying dependency between variables~\cite{xie2021logdp, lu2019LoPAD, paulheim2015decomposition}. Thus, dependency-based approach naturally leads to the conversion of the unsupervised anomaly detection problem to supervised learning problems. Given that there has been an extensive collection of feature selection and prediction techniques available, a meaningful and effective way for anomaly detection is to make use of off-the-shelf supervised techniques, instead of reinventing the wheel by developing new methods.
	
	Despite that research~\cite{xie2021logdp, lu2019LoPAD} has shown the promise of dependency-based anomaly detection, there are still certain research gaps in this area that need attention. Firstly, existing dependency-based methods represent only a fraction of a much larger potential combinations of supervised methods and scoring functions for dependency-based anomaly detection. There has been no work on summarizing the common procedure taken by the existing methods for establishing a framework to enable a systematic way of creating effective combinations of supervised learning and scoring techniques for dependency-based anomaly detection. Secondly, dependency-based methods are promising and practical, especially when off-the-shelf techniques are used. However, there is a lack of understanding on what choices of the off-the-shelf techniques would lead to better anomaly detection performance. Lastly, although the dependency-based approach is advantageous in providing meaningful interpretation in principle, there is a need for research on how to achieve this. 
	
	To address these gaps, this paper introduces a \textbf{Dep}endency-based \textbf{A}nomaly \textbf{D}etection framework (DepAD) to provide a general approach to dependency-based anomaly detection. For each phase of the DepAD framework, this paper analyzes what and how to utilize the off-the-shelf techniques in the context of anomaly detection. We systematically and empirically investigate the performance of different off-the-shelf techniques in each phase and their combinations to identify well-performing methods. We also examine the impact of different data characteristics on the performance of the DepAD methods. Lastly, we use a case study to demonstrate the interpretability of the DepAD framework. 
	
	In summary, the main contributions of this work are as follows:
	\begin{itemize}
		\item We propose a dependency-based anomaly detection framework, DepAD, to provide a general approach to dependency-based anomaly detection. DepAD offers a holistic approach to guide the development of dependency-based anomaly detection methods. DepAD is effective and adaptable, utilizing off-the-shelf techniques for diverse applications.
		
		\item We systematically and empirically study the performance of representative off-the-shelf techniques and their combinations in the DepAD framework. We identify two well-performing dependency-based methods. The two DepAD algorithms consistently outperform nine benchmark algorithms on 32 datasets.	
		
		\item We conduct a case study to demonstrate the capability of DepAD to provide insightful interpretations for detected anomalies.
	\end{itemize}
	
	The rest of the paper is organized as follows. In Section \ref{sec:related work}, we survey the related work. Section \ref{sec:framework} introduces the DepAD framework and presents the outline of the algorithms instantiated from DepAD. In Section \ref{sec:experiments}, we empirically study the performance of the DepAD methods and present the comparison of the proposed methods with state-of-the-art anomaly detection methods. Section \ref{sec:interpretation} explains how to interpret anomalies identified by DepAD methods using a case study. Section \ref{sec:conclusion} concludes the paper.
	
	\section{Related Work} \label{sec:related work}
	Various anomaly detection methods have been developed to leverage the distinctive characteristics of anomalies that deviate from the norm in some manner. The typical process of anomaly detection involves assuming a specific aspect in which anomalies are considered abnormal and then assessing the anomalousness of objects based on this aspect. In this section, we provide a concise overview of the proximity-based approach and then delve into existing dependency-based methods. Subsequently, we briefly review the subspace approach, emphasizing the similarities and differences between this approach and DepAD~\cite{thudumu2020comprehensive, aggarwal2016outlier}.
	
	\subsection{Proximity-based Approach}
	The proximity-based approach is mainstream in anomaly detection~\cite{yan2022structural, sarmadi2020novel, ramaswamy2000efficient, breunig2000lof}, and operates on the assumption that anomalies are objects that exhibit significant distance or sparsity in their neighborhood compared to other objects. The anomalousness of an object is determined by its proximity to neighboring objects. Proximity-based methods can be categorized into distance-based, density-based, and clustering-based methods. Distance-based methods utilize the distances between objects to assess anomalousness, such as the distance to the $k^{th}$ nearest neighbor~\cite{ angiulli2005outlier}. Density-based methods evaluate the anomalousness of an object based on the density of its neighborhood, exemplified by the Local Outlier Factor (LOF)~\cite{breunig2000lof} and its variants~\cite{tang2002enhancing, zhang2009new, kriegel2009loop}. Clustering-based methods assume that normal objects exhibit similarity and form clusters, while anomalies reside far away from the clusters or form small clusters.
	
	\subsection{Dependency-based Approach} \label{sec:existing methods}
	The dependency-based approach works under the assumption that anomalies deviate from the normal dependency among variables, and the extend of anomalousness is evaluated based on this deviation. While the proximity-based approach that focuses on relationships among objects, the dependency-based approach emphasizes on the relationships among variables. Dependency-based methods are comparatively less prevalent than proximity-based methods. This section offers an extensive review of the existing dependency-based methods.
	
	CFA~\cite{huang2003cross} is an anomaly detection method used in mobile ad hoc networking. It leverages correlations among variables to identify anomalies. For each variable, CFA uses other variables as predictors to estimate its expected value. The method first discretizes continuous values and then employs classification models (C4.5, RIPPER, and naive Bayes) to predict expected values. Two combination functions, average match count and average probability, are used for anomaly score generation, with the average probability showing a better performance in experiments. However, CFA's effectiveness with continuous data may be impacted by the discretization process it employs.
	
	FraC~\cite{noto2010anomaly,noto2012frac} is an anomaly detection method designed for continuous data, aiming to improve detection effectiveness. It uses an ensemble of three regression models (RBF kernel SVMs, linear kernel SVMs, and regression trees) to predict the expected values of variables based on their dependencies with other variables. Anomaly scores are computed as the sum of surprisals, which represent the log-loss of dependency deviations, over all variables.
	
	ALSO~\cite{paulheim2015decomposition} employs regression models (linear, Isotonic, and M5') to predict expected values based on the dependencies with other variables. Anomaly scores are calculated as the sum of weighted losses, squared dependency deviations weighted by the prediction accuracy of the regression model, across all variables.
	
	Both FraC and ALSO use all other variables as predictors, leading to high computational time and/or low accuracy in high-dimensional data. In contrast, COMBN~\cite{babbar2012mining} and LoPAD~\cite{lu2019LoPAD} propose using a subset of variables as predictors. COMBN uses Bayesian Networks (BN) to represent variable dependencies, using parents as predictors. However, this approach may miss anomalies resulting from non-parent variables and does not scale with high-dimensional data since it requires a complete BN.
	
	LoPAD addresses the limitations of COMBN by using the Markov Blanket (MB) of a variable as predictors. The MB contains all parents, children, and spouses of the variable. With this approach, LoPAD captures the complete dependency around the variable and accurately estimates its expected values. By utilizing regression trees with bagging for prediction and pruned summation for anomaly score computation, LoPAD effectively deals with high-dimensional data and achieves accurate anomaly detection.
	
	LogDP~\cite{xie2021logdp} is a semi-supervised log-based anomaly detection approach tailored for large-scale service-oriented systems troubleshooting. LogDP leverages the dependency relationships among log events and proximity among log sequences to identify anomalies in extensive unlabeled log data. By categorizing events, learning normal patterns, and using the MB as predictors with MLP regressors, LogDP achieves a superior performance compared to six state-of-the-art methods in experiments on real-world data.
	
	Recent research in the data-centric AI (DCAI) domain has leveraged anomaly detection techniques to identify out-of-distribution samples and data inconsistencies ~\cite{liang2022advances, huynh2024dagnosis, liu2020energy}. A representative method, DAGnosis~\cite{huynh2024dagnosis}, uses dependency-based approach to effectively detect and interpret inconsistencies. This method employs directed acyclic graphs (DAGs) to exploit data sparsity and independence among variables. For each variable, DAGnosis constructs a conformal prediction model, specifically a Conformalized Quantile Regression, to establish prediction intervals at a predefined significance level. If a feature of an object falls outside these prediction intervals, this object is flagged as an inconsistency. When building the conformal prediction model, the Markov blanket (MB) of each feature is used as predictors to estimate the feature value and to provide a explanation for the detected inconsistencies.
		
	In summary, these dependency-based methods leverage the relationships among variables to identify anomalies or inconsistencies in data. We outline and analyze the general workflow followed by these methods in depth in Section~\ref{sec:framework}.
	
	\subsection{Connection of DepAD with Subspace Anomaly Detection}
	Some readers may wonder what the differences are between DepAD and subspace anomaly detection approaches since both use a subset of variables for anomaly detection. We differentiate them in this subsection. To tackle the problem of anomaly detection in high-dimensional data, subspace anomaly detection methods, like those in~\cite{zimek2012survey, yu2018markov, kriegel2012outlier, thudumu2020comprehensive}, have been proposed to detect anomalies with proximity-based approach in a subspace containing a subset of variables. When determining subspaces, many methods (e.g.,~\cite{yu2018markov,kriegel2009outlier,kriegel2012outlier}) utilize the correlation of variables, and some (e.g.,~\cite{liu2008isolation,lazarevic2005feature}) randomly select variables. When evaluating anomalousness in each subspace, the criteria used in the proximity-based algorithms, such as LOF and kNN, are used. Comparing the subspace approach with DepAD, subspace anomaly detection methods are focused on searching for subsets of variables (subspaces) in which anomalous patterns can be more clearly identified than in the full space; whereas DepAD considers each variable and its predictors to assess the deviation of the observed value from the predicted value. Fundamentally, subspace anomaly detection methods make use of proximity in subspaces to detect anomalies; DepAD utilizes value deviation based on variable dependency to detect anomalies.
	
	\section{The DepAD Framework} \label{sec:framework}
	In this section, we introduce the DepAD framework. We begin with an overview of the framework and then proceed to explain each phase in detail. For each phase, we discuss its goal, key considerations, and the off-the-shelf techniques that can be utilized. Finally, we present the algorithm for instantiating the DepAD framework.
	
	We use an upper case letter, e.g., $X$ to denote a variable; a lower case letter, e.g., $x$ for a value of a variable; and a boldfaced upper case letter, e.g., $\textbf{X} = \{ X_1 , X_2, \cdots, X_m \}$ for a set of variables. We have reserved the letter $\textbf{D}$ for a data matrix of $n$ objects and $m$ variables $\{ X_1 , X_2, \cdots, X_m \}$ . We use $\textbf{x}_{i*}$ for the \textit{i}-th row vector (data point or object) of $\textbf{D}$, $\textbf{x}_{*j}$ for the \textit{j}-th column vector i.e., all values of variable $X_j$ in $\textbf{D}$, and $x_{ij}$ for the \textit{j}-th element in $\textbf{x}_{i*}$. For object $ \textbf{x}_{i*}$, its predicted value is denoted as $\hat{\textbf{x}}_{i*}$, and the predicted value of $x_{ij}$ is denoted as $\hat{x}_{ij}$. For a variable $X_j$, its set of relevant variables (i.e., predictors) is denoted as $\textbf{R}(X_j)$, and the value of $\textbf{R}(X_j)$ in the \textit{i}-th object in $\textbf{D}$ is represented as $\textbf{r}(x_{ij})$.
	
	\subsection{Overview of the DepAD Framework} 
	\label{subsec:framework}
\begin{figure*}[htbp]
		\centering
		\includegraphics[width=0.8\textwidth]{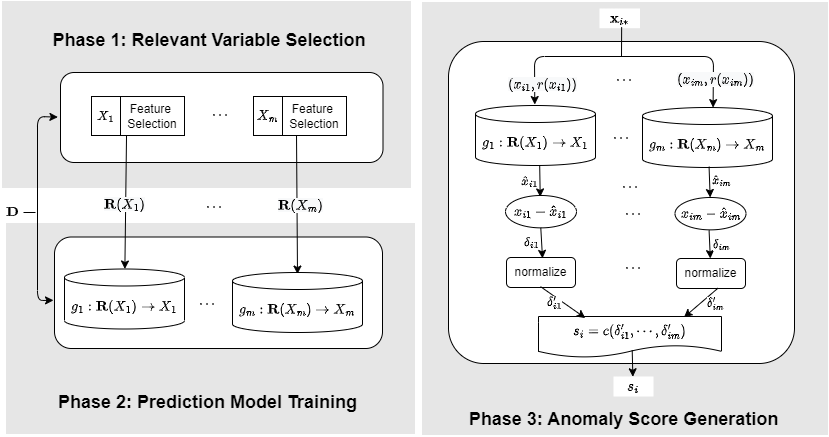}
		\caption{The DepAD framework}
		\label{fig:framework}
	\end{figure*}

	As depicted in Figure \ref{fig:framework}, DepAD consists of three phases: (1) relevant variable selection, (2) prediction model training, and (3) anomaly score generation.
	
	In Phase 1, for each variable $X_i$, we find the set of variables that are strongly related to $X_i$, called \emph{relevant variables} of $X_i$ and denoted as $\mathbf{R}(X_i)$, as the predictors for $ X_i $. Then, in Phase 2, a set of $m$ dependency prediction models is constructed, with one model for each variable. For example, for variable $X_i$, a model is trained by using $\mathbf{R}(X_i)$ as predictors and $X_i$ as the target variable. In Phase 3, given an object $\mathbf{x}_{i*}$, the expected values $\hat{x}_{ij}$ for each variable $X_j$ ($j \in {1, \ldots, m}$) are estimated using the learned prediction models from Phase 2 with values in $\mathbf{x}_{i*} \backslash x_{ij}$ ($\backslash$ indicates set difference) as input values obtaining from $\mathbf{D}$. The dependency deviations $\delta_{ij}$ are then calculated as the differences between the observed values $x_{ij}$ and their respective expected values $\hat{x}_{ij}$. This process generates a deviation vector, $\boldsymbol{\delta}_{i*}$, for $\mathbf{x}_{i*}$. Finally, the deviation vector $\boldsymbol{\delta}_{i*}$ is fed into a combination function $c$ to obtain the anomaly score, $s_i$, for the object $\textbf{x}_{i*}$. Anomalies are identified as the top-scored objects or those with scores exceeding a user-defined threshold.
	
	\subsection{Phase 1: Relevant Variable Selection} \label{sec:rel}
	The objective of this phase is to select the relevant variables, i.e., the predictor variables for each variable to create a prediction model for the variable (as the target variable of the model) using the selected variables. That is, for each variable $X_j \in \textbf{X}$, find a set of other variables $\textbf{R}(X_j) \subseteq \textbf{X} \backslash \{{X_j}\}$ that capture the dependency between $X_j$ and other variables in $\textbf{D}$. The selected relevant variables allow us to estimate the expected value of variable $X_j$ in an object $\textbf{x}_{i*}$, denoted as $\hat{x}_{ij}$, based on the values of its relevant variables, i.e.,$r(X_j) \in \mathbf{R}(X_j)$.
	
	This phase offers several advantages to DepAD. Firstly, relevant variable selection can eliminate redundant and irrelevant variables from the prediction models, reducing the risk of overfitting and enhancing prediction reliability. Secondly, it speeds up model training and enhances scalability, especially for high-dimensional data. Lastly, and notably, relevant variables facilitate the interpretation of detected anomalies, particularly in high-dimensional data.
	
	This phase can utilize off-the-shelf feature selection methods~\cite{yu2018unified,li2017feature} to identify the relevant variables. When choosing a feature selection method, the following factors should be considered: (1) The prediction models used in the prediction model training phase; (2) The interpretability of the selected variables; and (3) The efficiency of the feature selection method. In terms of interpretability (point 2), a small, non-redundant, and strongly related set of relevant variables is beneficial for interpreting detected anomalies.Lastly, different feature selection methods have varying computational costs and scalabilities based on the number of variables and objects. Thus, computational efficiency needs to be considered for different types of data.
	
	Feature selection methods fall into three categories, wrapper, filter and embedded methods~\cite{li2017feature}. Conceptually, methods from the three categories can all be utilized for relevant variable selection in DepAD. However, filter feature selection is preferred due to its high efficiency and independence of prediction models. 
	
	Among the filter feature selection methods, causal feature selection methods~\cite{yu2019causality,yu2018unified,guyon2007causal} are recommended choices as they identify the causal factors of a target variable, offering better interpretability. These methods select the parents and children (PC) or Markov blanket (MB) of a target variable in a Bayesian network (BN) as predictors for the target. The MB is considered an optimal choice for relevant variables as it contains the complete dependency information of a variable. Efficient methods exist for learning the PC or MB of a variable from data without learning a complete BN~\cite{yu2019causality, guyon2007causal}.
	
	Using all other variables as relevant variables, as done in some existing methods~\cite{noto2012frac,paulheim2015decomposition}, may reduce anomaly detection effectiveness and model efficiency, and hinder the interpretation of detected anomalies. 
	
	Table \ref{tbl:ph1_summary} provides a summary of commonly used filter feature selection methods that can be employed for relevant variable selection in the DepAD framework.
	\begin{table}[tbh]
		\caption{Summary of commonly used feature selection methods.}
		\label{tbl:ph1_summary}
		\centering
		\resizebox{\columnwidth}{!}{%
			\begin{threeparttable} 
				\begin{tabular}{lll}
					\toprule
					\textbf{Categories} & \textbf{Types} & \textbf{Methods} \\
					\midrule
					\multirow{2}{*}{Causal} & MB* & FBED~\cite{borboudakis2019forward}, HITON-MB~\cite{aliferis2003hiton}, IAMB~\cite{aragam2015concave}, Fast-IAMB~\cite{yaramakala2005speculative}, GSMB~\cite{margaritis2000bayesian} \\
					\cmidrule{2-3}
					& PC* & HITON-PC~\cite{aliferis2003hiton}, GET-PC~\cite{pena2005scalable} \\
					\midrule
					\multirow{4}{*}{Non-causal} & MI-based* & MI~\cite{battiti1994using}, NMI~\cite{estevez2009normalized}, MRMR~\cite{peng2005feature}, QPFS~\cite{guo2008gait}, SPEC~\cite{nguyen2014effective} \\
					\cmidrule{2-3}
					& Consistency-based & IEC~\cite{dash2003consistency}, IEPC~\cite{arauzo2008consistency} \\
					\cmidrule{2-3}
					& Dependency-based & DC~\cite{dodge2008coefficient}, BAHSIC~\cite{song2012feature}, FOHSIC~\cite{song2012feature} \\
					\cmidrule{2-3}
					& Distance-based & Relief~\cite{kira1992practical}, ReliefF~\cite{kononenko1994estimating}, RReliefF~\cite{robnik1997adaptation}\\
					\bottomrule
				\end{tabular}
				\begin{tablenotes}
					\small
					\item[*] MB: Markov Blanket; PC: Parent and Child; MI: Mutual Information.
				\end{tablenotes}
			\end{threeparttable} 
		}
	\end{table}
	
	\subsection{Phase 2: Prediction Model Training}
	In Phase 2, a set of prediction models is trained, one for each variable, using its relevant variables as predictors and itself as the target variable. The goal is to build a set of prediction models $\textbf{G} = \left\lbrace g_1, \cdots, g_m \right\rbrace$ using the input dataset $\textbf{D}$, where $g_j: \textbf{R}(X_j) \rightarrow X_j$. For an object $\textbf{x}_{i*}$, the expected value of variable $X_j$ in this object is $\hat{x}_{ij}=g_j(\textbf{r}(X_j))$. These prediction models are expected to produce small dependency deviations (i.e., the difference between the expected and observed values) for normal objects and large deviations for anomalies.
	
	When selecting a predictive model for DepAD, two important aspects need be considered.Firstly, the method is versatile and accurate. Versatility means that a method can be used for various data types and deal with different relationship types. Secondly, since the training set may contain anomalies, the prediction model should be robust to anomalies.
	
	Regression trees are favorable since they can handle both linear and non-linear relationships. To enhance prediction accuracy, bagging is recommended in combination with regression trees. Bagging involves training multiple regression trees with re-sampled data sets aggregating predictions of multiple trees. This technique reduces the impact of anomalies in several ways: 1) re-sampling potentially reduces the number of anomalies in the sets; 2) it can separate clustered anomalies, and thus mitigate teh risk of a prediction model overfitting anomalies; 3) averaging predictions from multiple trees reduces errors from individual trees. For greater robustness, some variants of tree-based regression models can also be utilized. For instance, a robust version of random forest in~\cite{roy2012robustness} uses the median instead of the mean for prediction aggregation. This idea has been applied to bagging CART trees, referred to as mCART.
	
	We generally discourage using linear regression models because they perform well only when the data follows a linear relationship. 
	
	Table \ref{tbl:ph2_summary} provides a summary of some commonly used prediction models, including classification and regression models, which can be employed as prediction models when instantiating the DepAD framework.
	
	\begin{table} [tbh] 
		\caption{Summary of commonly used prediction models.}
		\label{tbl:ph2_summary} 
		\centering
		\resizebox{\columnwidth}{!}{ %
			\begin{tabular}{p{2cm}p{7cm}}
				\toprule
				\textbf{Types} & \textbf{Techniques} \\
				\midrule	
				Tree-based & mCART~\cite{roy2012robustness}, CART~\cite{breiman1984classification}, C4.5~\cite{quinlan2014c4}, M5p~\cite{wang1996induction}, CHAID~\cite{kass1980exploratory} \\ 
				\midrule	
				Rule-based & RIPPER~\cite{cohen1995fast}, AQ~\cite{michalski1986multi}, ITRULE~\cite{smyth1992information}\\ 
				\midrule	
				Bayesian-based & Naive Bayes (NC)~\cite{langley1992analysis}, Robust NC~\cite{ramoni2001robust}, Bayesian Network~\cite{heckerman1997bayesian} \\
				\midrule	
				Linear regression & Genearal Linear (Linear)~\cite{seber2009multivariate}, Ridge~\cite{hoerl1970ridge}, Elastic Net~\cite{zou2005regularization}, Lasso~\cite{tibshirani1996regression}, MARS~\cite{friedman1991multivariate} \\ 
				\midrule	
				SVM-based & SVM classifier~\cite{boser1992training}, Support Vector Regression (SVR)~\cite{awad2015support} \\
				\bottomrule
			\end{tabular} 
		}
	\end{table} 
	
	\subsection{Phase 3: Anomaly Score Generation}
	In this phase, the goal is to generate anomaly scores from dependency deviations. For an object $\mathbf{x}_{i*}$, we first estimate its expected value $\hat{\mathbf{x}}_{i*}$ using the prediction model from Phase 2. Next, we compute the dependency deviation for each variable $X_j$ as $\delta_{ij} = \lvert x_{ij} - \hat{x}_{ij} \rvert$, and then normalize it as $\delta'_{ij}$. Finally, we use a combination function $c$ over the normalized dependency deviations of all variables to generate the anomaly score of $\mathbf{x}_{i*}$, denoted as $s_i$.
	
	Using normalized dependency deviations is critical for generating anomaly scores since the degree of anomalousness expressed by a dependency deviation is relative to other deviations. Hence, normalization is necessary.
	
	To normalize the dependency deviations, we can consider using transformation functions from ensemble anomaly detection methods, such as the Z-score normalization~\cite{nguyen2010mining} and the probability calibration~\cite{kriegel2011interpreting}. However, these methods may not be robust to the extreme values in deviations and could reduce the distinction between normal and abnormal objects. We recommend a more robust function, the robust Z-score, which is based on the traditional Z-score but uses median and average absolute deviation (AAD) instead of mean and standard deviation, respectively. The robust Z-score is computed as $\delta'_{ij} = \frac{\delta_{ij} - \tilde{\mu}_{\boldsymbol{\delta}_{*j}}}{\tilde{\sigma}_{\boldsymbol{\delta}_{*j}}}$, where $\tilde{\mu}_{\boldsymbol{\delta}_{*j}}$ and $\tilde{\sigma}_{\boldsymbol{\delta}_{*j}}$ are the median and AAD of the deviation vector $\boldsymbol{\delta}_{*j}$, respectively.
	
	For combining functions, we need to consider the dilution effect, where an anomaly with a few large deviations may be undetectable due to being outweighed by normal objects with numerous small deviations. To address this, we suggest using the pruned sum (PS) combination method, which only sums up deviations greater than a threshold $\eta$. Specifically, for an object $\mathbf{x}_{i*}$, its anomaly score computed by PS is $s_i = \sum_j^m I_{ij}\delta'_{ij}$, where $I_{ij}$ is 0 when $\delta'_{ij}$ is less than $\eta$, and 1 otherwise. This approach emphasizes larger deviations, potentially corresponding to anomalies, to improve detection effectiveness.
	
	In Table \ref{tbl:ph3_summary}, we provide a list of commonly used anomaly score generation methods (normalization and combination methods) and suggest using robust Z-score normalization and PS combination for DepAD.
	
	\begin{table} [tbh] 
		\caption{Summary of techniques for anomaly score generation.}
		\label{tbl:ph3_summary} 
		\centering
		\resizebox{1\columnwidth}{!}{ %
			\begin{tabular}{p{2cm}p{7cm}}
				\toprule
				\textbf{Types} & \textbf{Techniques} \\
				\midrule								
				Normalization & Z-score, robust Z-Score, HeDES~\cite{nguyen2010mining}, GS~\cite{kriegel2011interpreting} \\
				Combination & PS~\cite{aggarwal2015theoretical}, HeDES~\cite{nguyen2010mining}, Maximum (Max), Summation (Sum), Weighted Sum (WSum)\\
				\bottomrule
			\end{tabular} 		
		}
	\end{table} 
	
	\subsection{DepAD Algorithm Family and Complexity Analysis}
	The DepAD algorithm family, instantiated from the DepAD framework, follows the same procedure, outlined in Algorithm \ref{alg:depad}. The input to an algorithm is a dataset $\mathbf{D}$ and relevant parameters, and the output is the anomaly scores of all objects in $\textbf{D}$. User can then identify anomalies by top $l$ scored objects or objects whose scores exceed a predefined threshold.
	
	In Algorithm \ref{alg:depad}, the relevant variable selection is performed for each variable at Line 2, followed by the prediction model training at Line 3. The anomaly score generation phase occurs between Lines 5 and 14, where expected value estimation is carried out at Line 7, and the dependency deviations are computed at Line 8. Subsequently, the deviations are normalized at Line 10 and aggregated into anomaly scores at Line 13. Finally, the algorithm outputs the anomaly scores at Line 15.
	
	\begin{algorithm}
		\caption{The Outline of DepAD Algorithms} \label{alg:depad}
		\begin{algorithmic}
			\State \textbf{Input:} $\mathbf{D}$: a dataset with $n$ objects and a set of variables $\mathbf{X}=\{X_1, \cdots, X_m \}$ and relevant parameters 	
			\State \textbf{Output:} $\mathbf{s}$: anomaly scores of the objects in $\mathbf{D}$
		\end{algorithmic}
		
		\begin{algorithmic}[1]
			\For{ $\forall X_j \in \mathbf{X}, \ j \in \{1, \cdots, m \}$ } 
			\State get relevant variable of $X_j$, $\mathbf{R}(X_j)$ 
			\State train prediction model $g_j: X_j = g_j(\mathbf{R}(X_j))$ 
			\EndFor 
			
			\For{$\forall X_j \in \mathbf{X}, \ j \in \{1, \cdots, m \}$ } 
			\For{ $\forall \mathbf{x}_{i*} \in \mathbf{D},\ i \in \{1, \cdots, n \}$} 
			\State predict expected value of $x_{ij}$, i.e., $\hat{x}_{ij} = g_j(\textbf{\textit{x}}_{i{\mathbf{R}(X_j)}}) $
			\State $\delta_{ij} = \lvert x_{ij} - \hat{x}_{ij} \rvert$ 
			\EndFor		
			\State normalize the deviations on $ X_j $
			\EndFor 
			
			\For{$\forall \mathbf{x}_{i*}, \ i \in \{ 1, \cdots, n \}$ } 	
			\State compute anomaly score of $\textbf{\textit{x}}_{i*}$, $s_i = c(\boldsymbol{\delta}_{i*})$ 
			\EndFor 
			\State output $\mathbf{s} = \{ s_1, \cdots, s_n \}$
		\end{algorithmic}
	\end{algorithm} 
	
	A DepAD algorithm is denoted by combining the acronyms of the specific techniques used in each phase, as shown in Tables \ref{tbl:ph1_summary}, \ref{tbl:ph2_summary}, and \ref{tbl:ph3_summary}, separated by `-'. For instance, if FBED is employed in Phase 1, CART in Phase 2, and PS in Phase 3, the corresponding DepAD algorithm is named as FBED-CART-PS.
	
	The time complexity of a DepAD algorithm depends on the techniques chosen for the first two phases. Let's consider an example, IAMB-CART-RZPS. For a dataset with $n$ objects and $m$ variables, the time complexity of discovering MB using IAMB for $m$ variables is $O(m^2\lambda)$~\cite{aragam2015concave}, where $\lambda$ represents the average size of MBs. The complexity of building CART trees is $O(m\lambda n\log n)$~\cite{breiman1984classification}, and the complexity of anomaly score generation is $O(nm)$. Therefore, the overall complexity is $O(m^2\lambda) + O(m\lambda n\log n)$.
	
	\section{Evaluation} \label{sec:experiments}

	In this section, we aim to answer the following questions: 
	\begin{enumerate}
		\item How do the different techniques in a phase of DepAD affect the performance of DepAD algorithms? (Section \ref{sec:exp_steps})
		\item How the different combinations of the techniques in different phases affect the performance of DepAD algorithms? (Section \ref{sec:exp_combination})
		\item Compared with state-of-the-art anomaly detection methods, how is the performance of the DepAD algorithms? (Section \ref{sec:exp_benchmark})
		\item How do different data charateristics affect the performance of DepAD algorithms? (Section \ref{sec:sensitivity} );
		\item What is the efficiency of DepAD algorithms? (Section \ref{sec:exp_efficiency})
	\end{enumerate}
	
	\subsection{Datasets and Evaluation Metrics} \label{sec:exp_dataset}
	Thirty-two real-world datasets are used for the evaluation. These datasets cover diverse domains, e.g., spam detection, molecular bioactivity detection, and image object recognition, as shown in Table \ref{tbl:data}. The AID362, Backdoor, MNIST and caltech16 datasets are obtained from the Kaggle data repository~\cite{kaggle}. The Pima, WBC, Stamps, Ionosphere and Bank datasets are obtained from the anomaly detection data repository~\cite{campos2016evaluation}. The others are retrieved from the UCI data repository~\cite{UciData}. These datasets are often used in anomaly detection literature. 
	
	We follow the common process to obtain ground truth labels. When a dataset consists of two classes, we designate the majority class as the normal class and the minority class as the anomalous class. For datasets with multiple classes of imbalanced sizes, we select one or a few minority classes as anomaly class(es). 
	
	\begin{table*}[ht] \centering
		\caption{Summary of the 32 real-world datasets used in the experiments.} 
		\label{tbl:data}
		\resizebox{0.75\textwidth}{!}{
			\begin{tabular}{ lcl rcrc }
				\toprule
				\textbf{name} & \textbf{\#variables} & \textbf{class label}
				& \makecell{\textbf{normal} \\ \textbf{class}} 
				& \makecell{\textbf{\#objects in} \\ \textbf{normal class}} 
				& \makecell{\textbf{anomaly} \\ \textbf{class}} 
				& \makecell{\textbf{\#objects in} \\ \textbf{anomaly class}} \\
				\midrule
				Wilt & 6 & n,w & n & 4578 & w & 261 \\
				Pima & 9 & no,yes & no & 500 & yes & 268 \\
				WBC & 10 & no,yes & no & 444 & yes & 10 \\
				Stamps & 10 & no,yes & no & 309 & yes & 6 \\
				Glass & 10 & 2,1,7,3,5,6 & 2,1,7,3,5 & 205 & 6 & 9 \\
				Gamma & 11 & h,g & g & 12332 & h & 6688 \\
				PageBlocks & 11 & 1,2,5,4,3 & 1 & 4913 & 3 & 28 \\
				Wine & 12 & 6,5,7,8,4,3,9 & 6,5,7,8,4 & 4873 & 3,9 & 25 \\
				HeartDisease & 14 & 0,1,2,3,4 & 0,1,2,3 & 290 & 4 & 13\\
				Leaf & 16 & 1-36 & 1-35 & 330 & 36 & 10 \\
				Letter & 17 & A-Z & A & 789 & Z & 734 \\
				PenDigits & 17 & 0-9 & 1-9 & 9849 & 0 & 1143 \\
				Waveform & 22 & 2,1,0 & 2 & 1696 & 0 & 1657 \\
				Cardiotocography & 23 & 1,2,3 & 1 & 1655 & 3 & 176 \\
				Parkinson & 23 & 1,0 & 1 & 147 & 0 & 48 \\
				BreastCancer & 31 & B,M & B & 357 & M & 212 \\
				WBPC & 31 & N,R & N & 151 & R & 47 \\
				Ionosphere & 33 & no,yes & no & 225 & yes & 126 \\
				Biodegradation & 42 & NRB,RB & NRB & 699 & RB & 356 \\
				Bank & 52 & no,yes & no & 4000 & yes & 521 \\
				Spambase & 58 & 0,1 & 0 & 2788 & 1 & 1813 \\
				Libras & 91 & 1-15 & 2-15 & 336 & 1 & 24 \\
				Aid362 & 145 & inactive,active & inactive & 4219 & active & 60 \\
				Backdoor & 190 & normal,backdoor & normal & 56000 & backdoor & 1746 \\
				CalTech16 & 257 & 1-101 & 1 & 798 & 53 & 31 \\
				Arrhythmia & 275 & 1-16 & 1,2,10 & 339 & 14 & 4 \\
				Census & 386 & low,high & low & 44708 & high & 2683 \\
				Secom & 591 & -1,1 & -1 & 1463 & 1 & 104 \\
				MNIST & 785 & 0-9 & 7 & 1028 & 0 & 980 \\
				CalTech28 & 785 & 1-101 & 1 & 798 & 34 & 65 \\
				Fashion & 785 & 0-9 & 1 & 1000 & 0 & 1000 \\
				Ads & 1559 & nonad,ad & nonad & 2820 & ad & 459 \\
				\bottomrule
			\end{tabular}
		}
	\end{table*}
	
	We evaluate the performance of anomaly detection methods with two commonly used metrics: the Area under the Receiver Operating Characteristic Curve (ROC AUC)~\cite{liu2009encyclopedia} and Average Precision (AP)~\cite{campos2016evaluation}. ROC AUC measures the overall performance of the method, ranging from 0 to 1, where a value of 1 indicates perfect performance, and 0.5 indicates a random guess. AP, on the other hand, focuses on the ranks of anomalies and is defined as the average precision over the top-$l$ scored objects.
	
	For each dataset, we conduct repeated experiments to ensure robustness. If the ratio of anomalies to the total number of objects in the dataset is greater than $1\%$, we randomly sample $1\%$ of the total number of objects from the anomalous class as anomalies. This sampling process is repeated 20 times to obtain 20 datasets, and the experiment is conducted on each of these datasets. The average result over the 20 experiments is then reported. For datasets with a ratio of anomalies less than $1\%$, we conduct a single experiment using the dataset containing objects from both the normal and anomalous classes.
	
	To handle categorical variables, we convert them into numeric variables using 1-of-$\ell$ encoding~\cite{campos2016evaluation}. This ensures that all variables are represented numerically, allowing us to perform the necessary calculations during the evaluation process.
	
	\subsection{Evaluation of Techniques for Individual Phases of DepAD} \label{sec:exp_steps}
	In this section, we answer the first question, i.e., how do the different techniques in each individual phase affect the performance of DepAD algorithms?
	
	\subsubsection{Experiment Setting}
	To evaluate the performance of different techniques, as shown in Table \ref{tbl:dep_methods_125}, 5 representative techniques from different categories are chosen for each phase of DepAD, from which 125 algorithms in total are instantiated. Every time we evaluate a technique for a phase, we fix the techniques used in the other two phases. This gives us 25 results for the evaluated technique in the phase because the other two phases each have 5 different techniques used in the experiments. 
	
	\begin{table}[tbh]
		\caption{Representative techniques evaluated in our experiments.}
		\label{tbl:dep_methods_125} 
		\centering
		\resizebox{0.71\columnwidth}{!}{
			\begin{tabular}{cl}
				\toprule
				\textbf{Phase} & \textbf{Techniques} \\
				\midrule
				Phase 1 & FBED, HITON-PC, MI, IEPC, DC \\
				
				Phase 2 & CART, mCART, Linear, Ridge, Lasso \\
				
				Phase 3 & RZPS, PS, Sum, Max, GS \\
				\bottomrule
			\end{tabular}
		}
	\end{table}
	
	For Phase 1, five feature selection methods, including 2 causal and 3 non-causal methods, are used in our experiments. FBED and HITON-PC are causal feature selection techniques. FBED is used for MB (Markov Blanket) discovery and HITON-PC is for PC (Parents and Children) selection. MI, IEPC and DA are non-causal feature selection methods. MI is a mutual information-based feature selection method. IEPC and DC are consistency-based and dependency-based methods, respectively. For FBED and HITON-PC, we use their implementations in the \textit{CausalFS} toolbox~\cite{yu2019causality, yu2018unified}, and the significance level for the conditional independent tests is set to 0.01 for both of them. For MI, IEPC and DC, we use their implementations in the R package \textit{FSinR}~\cite{FSinR}. The slope thresholds in the three techniques for selecting features are all set to 0.8, as recommended by the package.
	
	For Phase 2, the prediction model training phase, we use the implementation of R package \textit{IPred}~\cite{ipred} for CART, and package \textit{glmnet}~\cite{glmnet} for Linear, Ridge and Lasso regression algorithms. Bagging is used with mCART and CART, where the number of trees is set to 25, the minimum number of a node to be split is 20, the minimum number of objects in a bucket is 7, and the complexity parameter is 0.003. For Lasso and Ridge, 10-fold cross validation is used to determine $\lambda$, the regularization parameter.
	
	For Phase 3, the anomaly score generation phase, we implement the techniques by ourselves in R. The threshold used in RZPS and PS is set to 0.
	
	\subsubsection{Performance of Relevant Variable Selection Techniques}
\begin{figure}[htbp] 
		\centering 
		\begin{subfigure}{0.48\columnwidth}
			\centering
			\includegraphics[width=0.99\columnwidth]{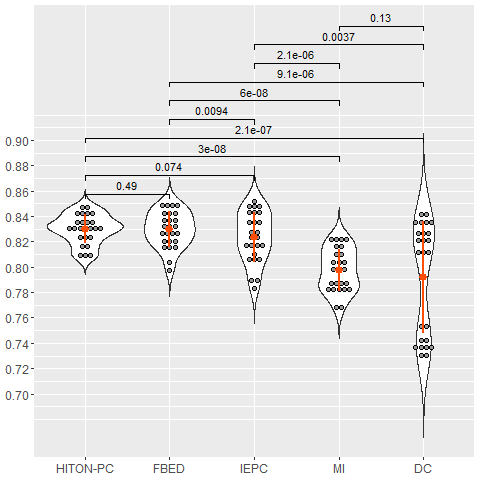}
			\caption{ROC AUC}
			\label{fig:rel_roc}
		\end{subfigure}
		\begin{subfigure}{0.48\columnwidth}
			\centering
			\includegraphics[width=0.99\columnwidth]{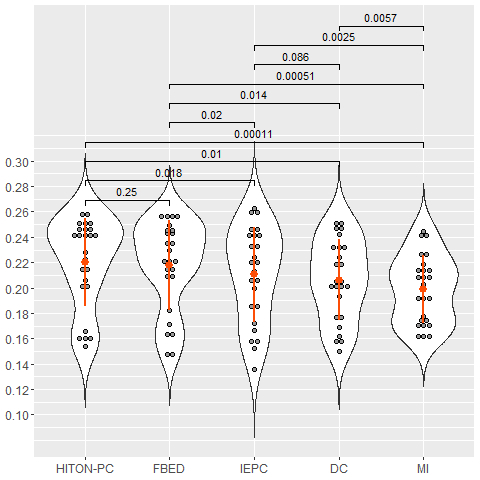}
			\caption{AP}
			\label{fig:rel_ap}
		\end{subfigure}
		\caption{Performance of relevant variable selection techniques. For each technique, 25 results averaged from 32 datasets are shown in black dots. The violin plot's outline indicates the Gaussian kernel density estimate of these results, with a red dot for the mean and lines indicating standard deviation. Techniques are pairwise compared using the Wilcoxon rank-sum test, with the left presumed superior; 	$p$-values are displayed above each pair.}
		\label{fig:rel} 
	\end{figure}

	The experimental results (ROC AUC and AP) of the five relevant variable selection techniques are shown in Figure \ref{fig:rel}. For each technique, its 25 results (each is the average results over the 32 datasets) are presented with a violin plot overlaid by a dot plot. For the dot plot, each black dot corresponds to a result. For the violin plot, the outline represents the density estimated using Gaussian kernel function of the 25 results. The red dot is the mean value, and the length of the red line on either side of the mean shows the standard deviation of the 25 results. The Wilcoxon rank-sum tests are pairwisely applied to the techniques, with the alternative hypothesis stating that the technique on the left is better than the one on the right. The $p$-values are shown at the top of each pair of techniques. In the following discussion, if the $p$-value is less than 0.05, the test result is significant.
	
	As shown in Figure \ref{fig:rel_roc}, the two causal feature selection techniques, HITON-PC and FBED, show better performance than the other three techniques. HITON-PC has the best average results, followed by FBED, IEPC, MI and DC. From the $p$-values shown in the figure, HITON-PC is significatly better than MI and DC, and FBED is significantly better than the three non-causal techniques. DC shows a much larger variance than other techniques, with a standard deviation of 0.044, while the standard deviations of other techniques range from 0.011 to 0.017.
	
	
	Regarding AP, HITON-PC and FBED exhibit significantly better performance than the other three techniques, as depicted in Figure \ref{fig:rel_ap}. Notably, the results of AP generally display larger variances than those of ROC AUC, which indicates the unstable performance measuring with AP. 
	
	Table \ref{tbl:rel_size} presents the reduction rates achieved by each of the five techniques. The reduction rate is computed as 1 minus the ratio of the number of relevant variables selected to the total number of variables in a dataset. The results reveal substantial variations in reduction rates among the different techniques for the same dataset. For instance, for the dataset Libras, the reduction rate achieved by IEPC is 2.2\%, while the other techniques achieve rates below 93\%. On average, HITON-PC exhibits the highest reduction rate of 84.61\%, while IEPC shows the lowest reduction rate at 40.28\%. FBED, DC, and MI achieve relatively similar reduction rates, hovering around 76\%. Notably, FBED and HITON-PC display a similar trend, with HITON-PC consistently achieving a higher reduction rate than FBED due to the PC set of a variable being a subset of its Markov blanket. 
	
	Compared to other methods, IEPC exhibits a notably lower reduction rate, which, we believe, contributes to its unstable performance. The experimental results in Figure \ref{fig:rel} indicate that when considering only linear prediction models, IEPC performs better with regularization techniques such as LASSO and Ridge, as opposed to general linear regression without regularization. This observation suggests the possibility of irrelevant or redundant variables being included in the set of relevant variables selected by IEPC. 
	
	\begin{table}[htbp] \centering
		\caption{Average reduction rate of the five feature selection techniques.} 
		\renewcommand{\arraystretch}{0.9}
		\label{tbl:rel_size}
		\resizebox{1\columnwidth}{!}{
			\begin{tabular}{ lc rrrrr }
				\toprule 
				\textbf{Name} & \textbf{\#Variables} & \textbf{FBED} & \textbf{HITON-PC} & \textbf{IEPC} & \textbf{DC} & \textbf{MI} \\ 
				\midrule
				Wilt & 6 & 53.30\% & 56.30\% & 33.30\% & 60\% & 53.30\% \\
				Pima & 9 & 69.70\% & 75.20\% & 23.30\% & 88.90\% & 45.20\% \\
				WBC & 10 & 71.80\% & 71.80\% & 36.20\% & 36\% & 37.20\% \\
				Stamps & 10 & 70.60\% & 75\% & 29\% & 29\% & 73\% \\
				Glass & 10 & 53.80\% & 68.60\% & 38\% & 83\% & 54.50\% \\
				Gamma & 11 & 47.70\% & 62.30\% & 18.20\% & 42.70\% & 60.70\% \\
				PageBlocks & 11 & 38.20\% & 63.60\% & 26.40\% & 89.10\% & 71.80\% \\
				Wine & 12 & 42.50\% & 61.70\% & 16.70\% & 90.80\% & 64.20\% \\
				HeartDisease & 14 & 83.90\% & 84.20\% & 64.80\% & 92.80\% & 33.90\% \\
				Leaf & 16 & 72.10\% & 82.20\% & 24.40\% & 24.40\% & 85.60\% \\
				Letter & 17 & 55.40\% & 77.40\% & 17.10\% & 89.70\% & 92.60\% \\
				PenDigits & 17 & 25.80\% & 69.60\% & 27.60\% & 92.10\% & 77.60\% \\
				Waveform & 22 & 77.30\% & 77.50\% & 9.10\% & 74.10\% & 35.30\% \\
				Cardiotocography & 23 & 67\% & 82.60\% & 47.40\% & 86.20\% & 64.90\% \\
				Parkinson & 23 & 83.50\% & 89.70\% & 13\% & 39.70\% & 87.60\% \\
				BreastCancer & 31 & 76.50\% & 88.10\% & 6.50\% & 49.60\% & 94.20\% \\
				WBPC & 31 & 83.50\% & 90.60\% & 6.50\% & 22.10\% & 95\% \\
				Ionosphere & 33 & 80.90\% & 90.50\% & 11.80\% & 80.50\% & 95.30\% \\
				Biodegradation & 42 & 88.90\% & 80.40\% & 83.40\% & 64\% & 80.30\% \\
				Bank & 52 & 82.10\% & 88.30\% & 72.90\% & 94.50\% & 68.40\% \\
				Spambase & 58 & 89.90\% & 92.80\% & 67.70\% & 95.30\% & 58.10\% \\
				Libras & 91 & 93.20\% & 95.60\% & 2.20\% & 98.90\% & 97.60\% \\
				Aid362 & 145 & 82.80\% & 94.30\% & 50.70\% & 82.30\% & 99.20\% \\
				Backdoor & 190 & 93.90\% & 98.20\% & 36.20\% & 38.30\% & 42\% \\
				CalTech16 & 257 & 95.20\% & 98\% & 78.50\% & 98.10\% & 99.60\% \\
				Arrhythmia & 275 & 96.80\% & 98.50\% & 57.10\% & 98.20\% & 98.90\% \\
				Census & 386 & 94.10\% & 97.30\% & 44.70\% & 94.20\% & 95.70\% \\
				Secom & 591 & 98.10\% & 99.30\% & 36.20\% & 95.40\% & 99.10\% \\
				MNIST & 785 & 98.10\% & 99.30\% & 76.60\% & 99.40\% & 99.90\% \\
				calTech28 & 785 & 98.20\% & 99.40\% & 92\% & 99.80\% & 99.90\% \\
				Fashion & 785 & 97.20\% & 99.40\% & 51.50\% & 95.40\% & 99.70\% \\
				Ads & 1559 & 99.50\% & 99.70\% & 89.90\% & 99.70\% & 99.50\% \\
				\midrule
				Average & 197 & 76.92\% & 84.61\% & 40.28\% & 75.76\% & 76.87\% \\	
				\bottomrule
			\end{tabular}
		}
	\end{table}
	
	In conclusion, the relevant variable selection phase of the DepAD framework is crucial for identifying optimal predictors for the target variable in anomaly detection. Striking a balance between selecting too many or too few variables is essential for maintaining prediction accuracy. When the ground-truth relevant variable set is unavailable, the Markov blanket (MB) represents a theoretically optimal choice. Our experiments have further validated that HITON-PC and FBED outperform the other techniques, and achieve superior results in both ROC AUC and AP and the highest variable reduction rates. 
	
	\subsubsection{Performance of Different Prediction Models}
\begin{figure}[htbp] 
		\centering 
		\begin{subfigure}{0.48\columnwidth}
			\centering
			\includegraphics[width=0.99\columnwidth]{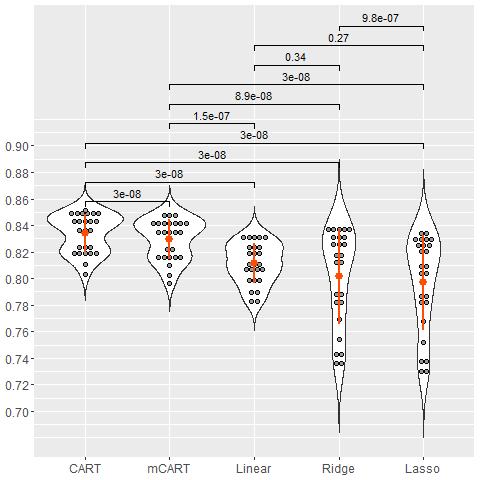}
			\caption{ROC AUC}
			\label{fig:pred_roc}
		\end{subfigure}
		\begin{subfigure}{0.48\columnwidth}
			\centering
			\includegraphics[width=0.99\columnwidth]{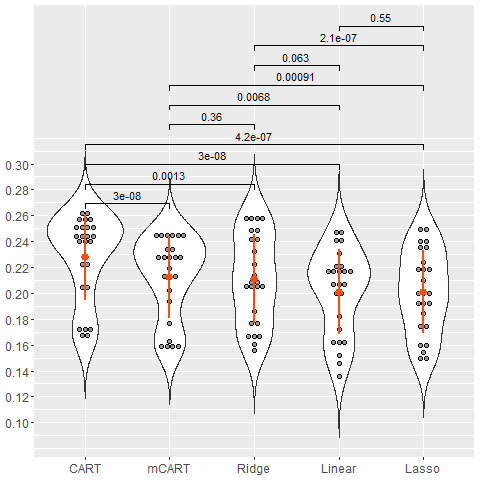}
			\caption{AP}
			\label{fig:pred_ap}
		\end{subfigure}
		\caption{Performance of different prediction models. For each technique, 25 results averaged from 32 datasets are shown in black dots. The violin plot's outline indicates the Gaussian kernel density estimate of these results, with a red dot for the mean and lines indicating standard deviation. Techniques are pairwise compared using the Wilcoxon rank-sum test, with the left presumed superior; $p$-values are displayed above each pair. }
		\label{fig:pred} 
	\end{figure}

	Concerning ROC AUC, the tree-based algorithms, CART and mCART, demonstrate superior performance compared to all the linear regression algorithms, as illustrated in Figure \ref{fig:pred_roc}. The $p$-values in the figure reveal that each regression algorithm (excluding Lasso) exhibits significant improvement over the algorithms to its right. Notably, Ridge and Lasso display unstable results with a standard deviation of 0.036, while the standard deviations of the other regression algorithms range from 0.014 to 0.015. 
	
	In terms of AP, CART achieves the highest average results, followed by mCART, Ridge, Linear, and Lasso, as depicted in Figure \ref{fig:pred_ap}. CART significantly outperforms all other techniques, and mCART demonstrates superiority over Linear and Lasso. Additionally, Ridge is significantly better than Lasso. Both Linear and Lasso exhibit the poorest performance. The variances of the five algorithms are comparable, ranging from 0.031 to 0.034.
	
	In summary, the two tree-based models, CART and mCART, demonstrate superior performance in both ROC AUC and AP. On the contrary, Lasso yields the lowest results for both metrics.
	
	\subsubsection{Performance of Anomaly Score Generation Techniques}
\begin{figure}[htbp] 
		\centering 
		\begin{subfigure}{0.48\columnwidth}
			\centering
			\includegraphics[width=0.99\columnwidth]{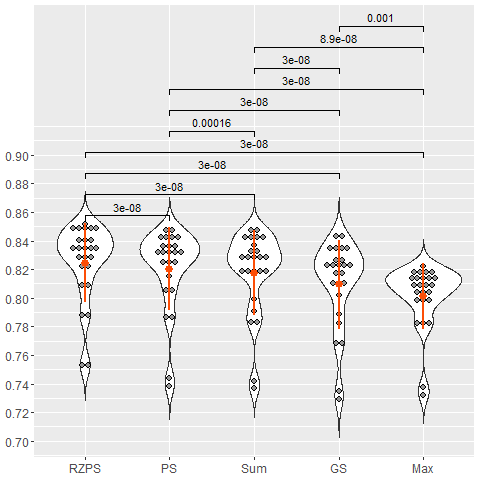}
			\caption{ROC AUC}
			\label{fig:comb_roc}
		\end{subfigure}
		\begin{subfigure}{0.48\columnwidth}
			\centering
			\includegraphics[width=0.99\columnwidth]{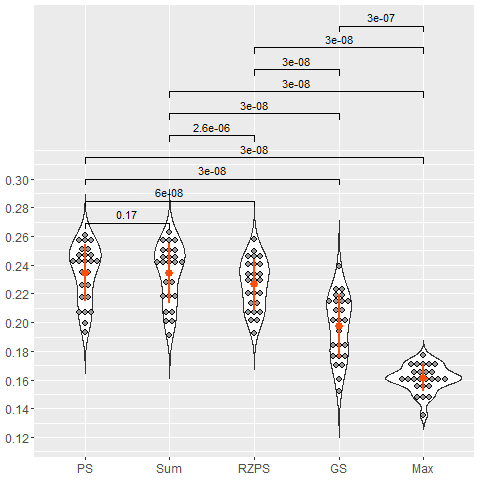}
			\caption{AP}
			\label{fig:comb_ap}
		\end{subfigure}
		\caption{Performance of anomaly score generation techniques. For each technique, 25 results averaged from 32 datasets are shown in black dots. The violin plot's outline indicates the Gaussian kernel density estimated from these results, with a red dot for the mean and lines indicating standard deviation. Techniques are pairwise compared using the Wilcoxon rank-sum test, with the left presumed superior; $p$-values are displayed above each pair. }
		\label{fig:comb} 
	\end{figure}

	The results of the five scoring methods are presented in Figure \ref{fig:comb}. In terms of ROC AUC, RZPS achieves the highest average performance, followed by PS, Sum, GS, and Max. As illustrated in Figure \ref{fig:comb_roc}, each technique shows significant improvement compared to the techniques to its right, as indicated by the $p$-values in the figure. The standard deviations of the results for all five techniques are similar, ranging from 0.023 to 0.031.
	
	Upon examining the violin plots of the five techniques in Figure \ref{fig:comb}, we observe more consistent shapes compared to the violin plots in Figures \ref{fig:rel} and \ref{fig:pred}. Most dots in Figure \ref{fig:comb} are located in the upper part, a few dots are located in the middle part and low parts. Each dot represents the performance of a combination of the other two phases. The consistent shapes in Figure \ref{fig:comb} suggest that the anomaly score generation techniques do not significantly alter the distribution of the 25 combinations. This implies that the choices of methods for relevant variable selection and prediction models have a more significant impact on the final performance than the choice of scoring techniques. Notably, the two dots at the bottom of the violin plots in Figure \ref{fig:comb} correspond to the combinations of DC with Lasso and Ridge, respectively, which will be further discussed in Section \ref{sec:exp_combination}.
	
	With respect to AP, PS and Sum achieve the best results, followed by RZPS, GS and Max, as shown in Figure \ref{fig:comb_ap}. Each technique is significantly better than the techniques to its right except for PS when compared with Sum. The results of Max are significantly worse than the results of the other techniques. The standard deviations of the results of the five techniques are similar, around 0.02, except Max, which has a slightly lower standard deviation of 0.009.
	
	In summary, RZPS has the best performance in terms of ROC AUC, while PS and Sum produce the best results in terms of AP. GS and Max show the worst performance, as indicated by both ROC AUC and AP.
	
	\subsection{Evaluation of Different Combinations of Techniques} \label{sec:exp_combination}
	In this subsection, we aim to address the second question of our study: How do different combinations of techniques in different phases of DepAD impact the overall performance of DepAD algorithms? Through our analysis, we identify specific combinations that may lead to poor results and propose combinations of techniques that can achieve better performance.
	
\begin{figure}[tbp] 
		\centering 
		\begin{subfigure}{0.99\columnwidth}
			\centering
			\includegraphics[width=0.99\columnwidth]{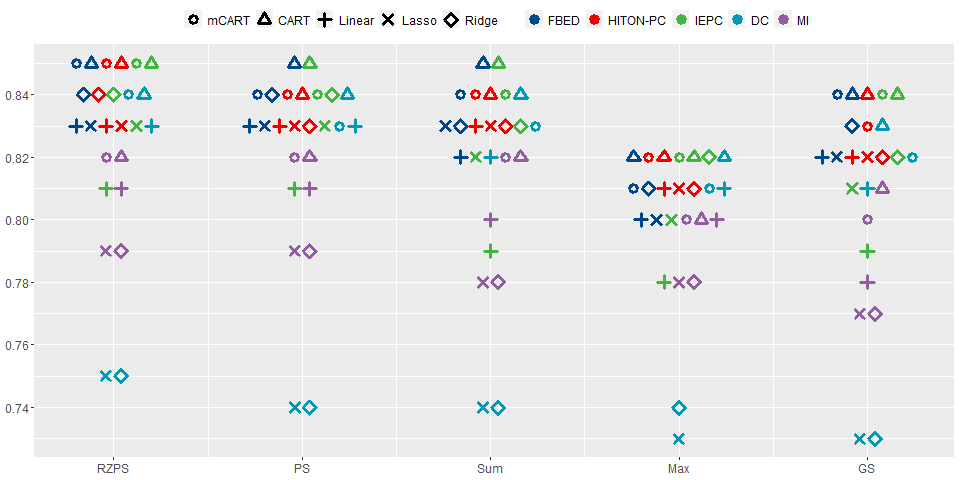}
			\caption{Results of the 125 DepAD algorithms in terms of ROC AUC}
			\label{fig:eva_combine_roc}
		\end{subfigure}
		\begin{subfigure}{0.99\columnwidth}
			\centering
			\includegraphics[width=0.99\columnwidth]{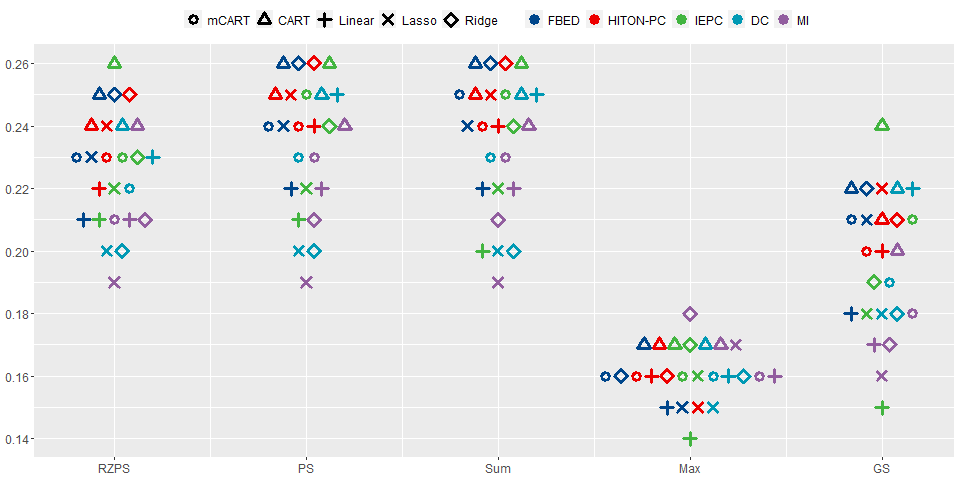}
			\caption{Results of the 125 DepAD algorithms in terms of AP}
			\label{fig:eva_combine_ap}
		\end{subfigure} 
		\caption{Results (ROC AUC and AP) of the 125 DepAD algorithms. Each sub-figure uses different colors for variable selection techniques and different shapes for prediction models, as shown at the top of the sub-figure. Results are grouped by the techniques used in the anomaly score generation phase.}
		\label{fig:eva_combination} 
	\end{figure}

	The results of the 125 DepAD algorithms are presented in Figure \ref{fig:eva_combination}, which consists of two sub-figures representing ROC AUC and AP, respectively. 
	
	Regarding ROC AUC, ten methods achieved the best results of 0.85. These methods resulted from combinations of Phase 1 techniques, including FBED, HITON-PC, and IEPC; Phase 2 techniques, specifically CART and mCART; and Phase 3 techniques, which involved RZPS, PS, and Sum, as depicted in Figure \ref{fig:eva_combine_roc}. Our analysis also revealed some specific observations: (1) When HITON-PC is utilized, the best results are attained only when combined with RZPS, not with PS or Sum; (2) When using PS or Sum as scoring techniques, the best results are achieved by combining them with CART as the prediction model; (3) If mCART is employed as the prediction model, the best results are obtained by combining it with RZPS. In summary, algorithms that utilize the combinations of FBED, CART, RZPS, PS, and Sum consistently achieve the best results in the experiments.
	
	Conversely, algorithms that contain combinations of DC with Lasso or Ridge demonstrate the worst performance. Additionally, algorithms using MI as the relevant variable selection technique generally show inferior results, regardless of the techniques used in the other two phases. When using IEPC for relevant variable selection, combining it with the Linear model yields much poorer results compared to combining IEPC with other prediction models.
	
	Regarding AP, the best-performing methods are derived from the following combinations: Phase 1 techniques include IEPC, FBED, and HITON-PC; Phase 2 techniques involve CART and Ridge; and Phase 3 techniques consist of PS, Sum, and RZPS, as illustrated in Figure \ref{fig:eva_combine_ap}. Further observations include: (1) When Ridge is used as the prediction model, the best results are achieved only when combined with PS or Sum; (2) If HITON-PC is adopted, the best results are obtained only when combined with Ridge; (3) When FBED is used, the best results are achieved only when combined with PS or Sum.
	
	Conversely, the results from the combination of IEPC with Ridge exhibit much lower performance compared to other combinations. Methods using Max as the scoring technique yield the worst results. Additionally, the results with MI as the relevant variable selection technique are generally inferior, regardless of the techniques used in the other two phases. Therefore, in terms of AP, IEPC-CART-PS, FBED-CART-PS, FBED-CART-Sum, and IEPC-CART-Sum show better performance.
	
	In summary, the DepAD methods FBED-CART-RZPS, FBED-CART-PS, and FBED-CART-Sum generally demonstrate good performance in terms of ROC AUC. Among them, FBED-CART-PS and FBED-CART-Sum are considered good choices as they exhibit favorable performance in both ROC AUC and AP. It is noteworthy that FBED-CART-PS is the same algorithm proposed in~\cite{lu2019LoPAD}.
	
	\subsection{Comparison with Benchmark Methods} \label{sec:exp_benchmark}
	In the subsection, we answer the question, i.e., compared with state-of-the-art anomaly detection methods, how is the performance of the instantiated DepAD algorithms? We choose the two DepAD algorithms, FBED-CART-PS and FBED-CART-Sum, to compare them with the nine state-of-the-art anomaly detection methods shown in Table \ref{tbl:benchmark}, including seven proximity-based methods and two dependency-based methods. The settings of these methods can be found in Table \ref{tbl:benchmark}.
	
	It is worth noting that the key difference between the two DepAD methods (FBED-CART-PS and FBED-CAR-Sum) and ALSO lies in their relevant variable selection phase. The two DepAD methods learn and use the MB of a variable as its relevant variables, while ALSO, for each variable, uses all other variables as the variable's relevant variables. Thus, the evaluation demonstrates the impact of the relevant variable selection phase.
	
	\begin{table}[htbp] \centering
		\caption{Summary of benchmark methods and parameter settings.} 
		\label{tbl:benchmark}
		\resizebox{1\columnwidth}{!}{
			\begin{threeparttable} 
				\begin{tabular}{ ll }
					\toprule
					\textbf{benchmark methods} & \textbf{parameter setting} \tnote{a} \\
					\midrule
					LOF~\cite{breunig2000lof} & the number of nearest neighbors is set to 10 \\
					wkNN~\cite{angiulli2005outlier} & the number of nearest neighbors is set to 10 \\
					FastABOD~\cite{kriegel2008angle} & the number of nearest neighbors is set to 10 \\
					iForest~\cite{liu2008isolation} & the number of trees is set to 100 without sub-sampling \\
					MBOM~\cite{yu2018markov} & the number of nearest neighbors is set to 10 \\
					SOD~\cite{kriegel2009outlier} & the number of shared nearest neighbors is 10\\
					OCSVM~\cite{scholkopf2001estimating} & RBF kernel; $\gamma$: $\{0.01, 0.03, 0.05, 0.07, 0.09\}$ \tnote{b} \\
					ALSO~\cite{paulheim2015decomposition} & M5' regression is used as the prediction model \\
					COMBN~\cite{babbar2012mining} & BN is learned using R \textit{bnlearn} package \\
					\bottomrule
				\end{tabular}
				\begin{tablenotes}\footnotesize
					\item[a] We adopt the commonly used or recommended parameters used in the original papers.
					\item[b] OCSVM: One-class SVM. As OCSVM is sensitive to the kernel coefficient setting $\gamma$, experiments conducted with $\gamma$ are set to 0.01, 0.03, 0.05, 0.07 and 0.09, then the best results are reported.
				\end{tablenotes}
			\end{threeparttable} 
		}
	\end{table} 
	
	For LOF, iForest, FastABOD, OCSVM and SOD, we use the implementations in the \textit{dbscan}~\cite{dbscan} R package, \textit{IsolationForest}~\cite{iforest} R package, \textit{abodOutlier}~\cite{abod} R package, \textit{e1071}~\cite{e1071} R package and \textit{HighDimOut}~\cite{HighDimOut} R package respectively. MBOM, ALSO and COMBN are implemented by ourselves based on the \textit{bnlearn}~\cite{bnlearnpackage} R package. All the methods are implemented using R, and the experiments are conducted on a high-performance computer cluster with 32 CPU cores and 128G memory. 
	
	In the experiments, if a method is unable to produce a result within four hours, we stop the experiments. The stopped methods and data sets include 1) FastABOD and SOD on datasets Backdoor and Census; 2) ALSO on datasets Backdoor, CalTech16, Census, Secom, MNIST, CalTech28, Fashion and Ads; 3) COMBN on datasets Backdoor, CalTech16, Arrhythmia, Census, Secom, MNIST, CalTech28, Fashion and Ads.
	
\begin{figure}[htbp] 
		\centering 
		\includegraphics[width=0.95\columnwidth]{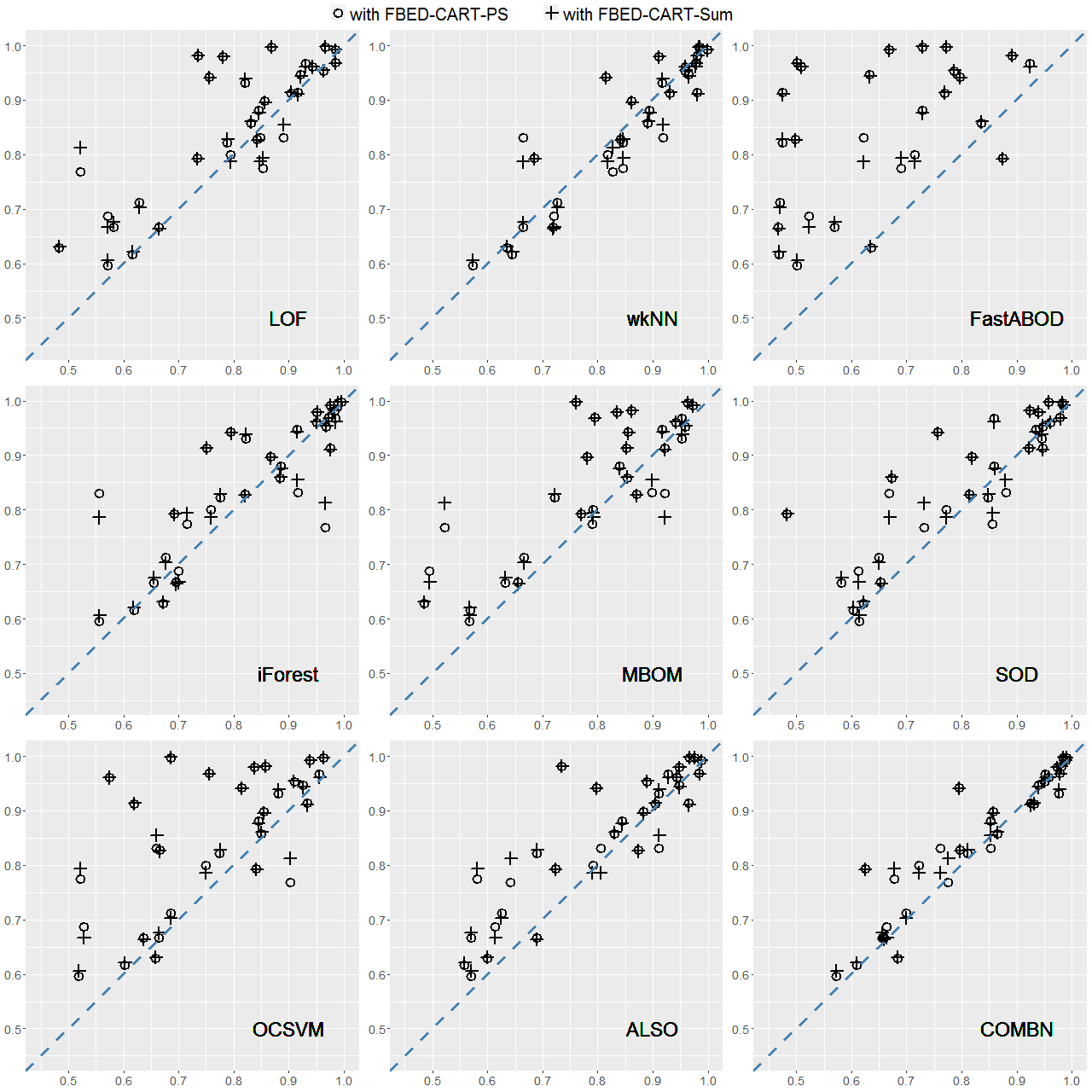}
		\caption{Comparison of two DepAD algorithms, FBED-CART-PS and FBED-CART-Sum, with benchmark methods in terms of ROC AUC. The X axis stands for the ROC AUC of a comparison method, and the Y axis represents the ROC AUC of FBED-CART-PS (circle) or FBED-CART-Sum (plus). A dot (or plus) represents a comparison of FBED-CART-PS (or FBED-CART-Sum) with a method named in each sub figure. A dot or plus falling in the top left of the diagonal line indicates FBED-CART-PS or FBED-CART-Sum performs better than the comparison method.}
		\label{fig:ben_roc}
	\end{figure}

\begin{figure}[htbp] 
		\centering
		\includegraphics[width=0.95\columnwidth]{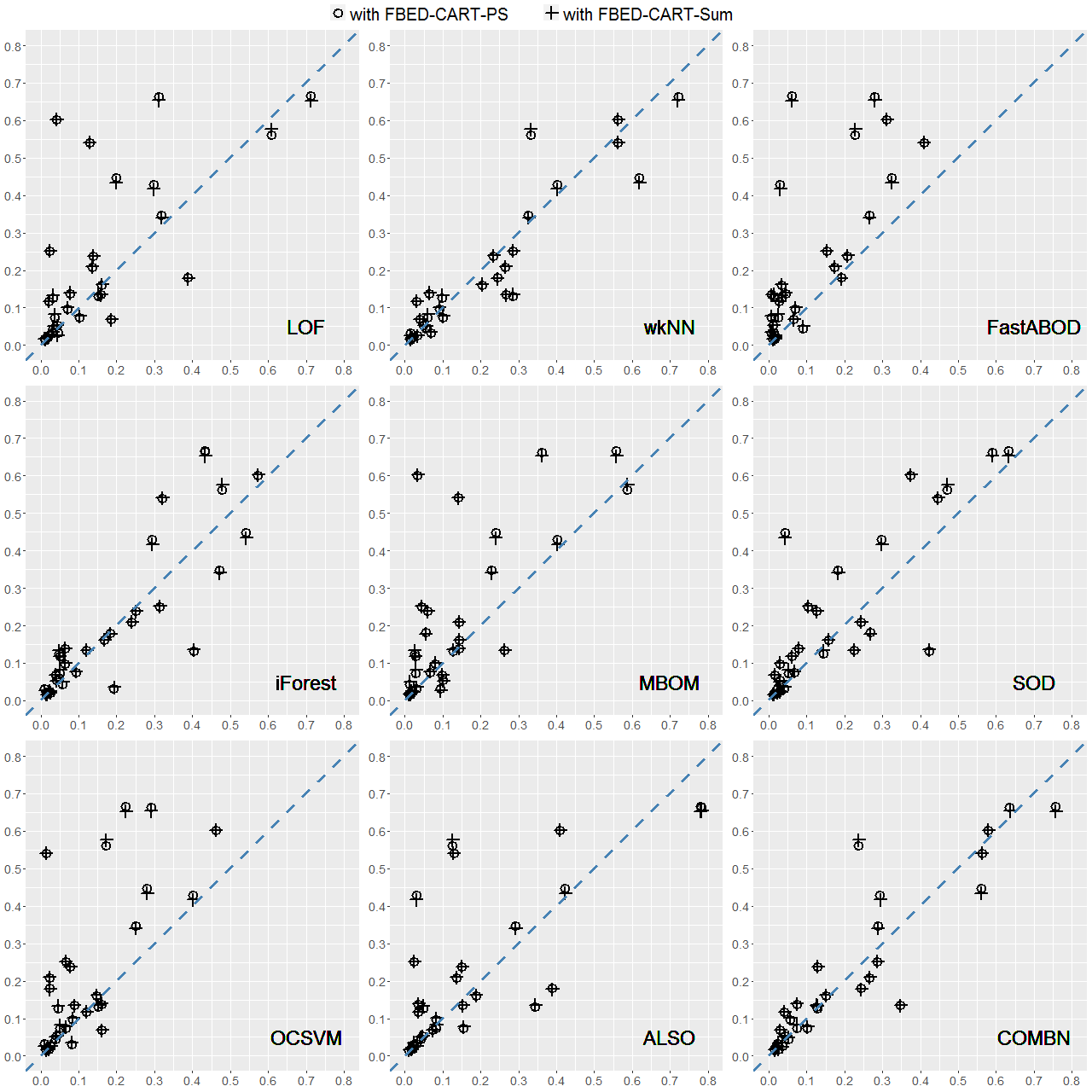}
		\caption{Comparison of two DepAD algorithms, FBED-CART-PS and FBED-CART-Sum, with benchmark methods in terms of AP. The X axis stands for the AP of a comparison method, and the Y axis represents the AP of FBED-CART-PS (circle) or FBED-CART-Sum (plus). A dot (or plus) represents a comparison of FBED-CART-PS (or FBED-CART-Sum) with a method named in each sub figure. A dot or plus falling in the top left of the diagonal line indicates FBED-CART-PS or FBED-CART-Sum performs better than the comparison method.}
		\label{fig:ben_ap}
	\end{figure}

	The comparison results are shown in Figures \ref{fig:ben_roc} (ROC AUC) and \ref{fig:ben_ap} (AP), where each sub-figure corresponds to the comparison results between the two DepAD algorithms with one benchmark method. In a sub-figure, the more circles or pluses sitting above the diagonal line, the more datasets on which the DepAD algorithm outperforms the comparison method. 
	
	Wilcoxon signed ranks tests are conducted on the results of each of the two DepAD algorithms, i.e., FBED-CART-PS and FBED-CART-Sum, pairwise with each of the nine benchmark methods. The alternative hypothesis is that a DepAD algorithm is better than the comparison method. The $p$-values are shown in Table \ref{tbl:p value}, where * indicates that the $p$-value is less than 0.05.
	
	\begin{table*}[htbp] \centering
		\renewcommand{\arraystretch}{0.8}
		\caption{$p$-values of Wilcoxon Signed Ranks Test on DepAD algorithms paired with the benchmark methods.}
		\label{tbl:p value}
		\resizebox{0.8\textwidth}{!}{
			\begin{threeparttable} 
				\begin{tabular}{ ll rrrrrrrrr }
					\toprule
					\textbf{Metrics} & \textbf{Methods} & \textbf{LOF} & \textbf{wkNN} & \textbf{FastABOD} & \textbf{iForest} & \textbf{MBOM} & \textbf{SOD} & \textbf{OCSVM} & \textbf{ALSO} & \textbf{COMBN} \\
					\midrule 
					\multirow{2}{*}{ROC AUC} & FBED-CART-PS & 0\tnote{*} & 0.892 & 0\tnote{*} & 0.062 & 0\tnote{*} & 0.002\tnote{*} & 0\tnote{*} & 0\tnote{*} & 0.007\tnote{*} \\
					& FBED-CART-Sum & 0\tnote{*} & 0.848 & 0\tnote{*} & 0.072 & 0\tnote{*} & 0.001\tnote{*} & 0\tnote{*} & 0\tnote{*} & 0.002\tnote{*} \\
					\midrule
					\multirow{2}{*}{AP} & FBED-CART-PS & 0.004\tnote{*} & 0.57 & 0\tnote{*} & 0.077 & 0\tnote{*} & 0\tnote{*} & 0.001\tnote{*} & 0.025\tnote{*} & 0.112 \\
					& FBED-CART-Sum & 0.003\tnote{*} & 0.54 & 0\tnote{*} & 0.067 & 0\tnote{*} & 0.001\tnote{*} & 0\tnote{*} & 0.02\tnote{*} & 0.086 \\
					\bottomrule
				\end{tabular}
				\begin{tablenotes}
					\item[*] The $p$-value is less than 0.05.
				\end{tablenotes}
			\end{threeparttable} 
		}
	\end{table*}
	
	According to Figure \ref{fig:ben_roc} and Table \ref{tbl:p value}, the two DepAD algorithms are significantly better than all benchmark methods except for wkNN and iForest in terms of ROC AUC . With wkNN, the results are similar. With iForest, the $p$-values are very close to 0.05. In terms of AP, the two DepAD algorithms yield significantly better results than all benchmark methods except for wKNN, iForest and COMBN, as shown in Figure \ref{fig:ben_ap} and Table \ref{tbl:p value}. With wkNN, the $p$-value is around 0.5, which shows a similar performance. The $p$-values with iForest and COMBN are close to 0.05. Furthermore, the two DepAD methods significantly outperform ALSO, and this is attributed to the inclusion of the relevant variable selection. In summary, the two DepAD algorithms outperform most of the benchmark methods, including both proximity-based methods and existing dependency-based methods.
	
	\paragraph{\textbf{A Further Analysis of DepAD Algorithms and wkNN Performance}}
	As FBED-CART-PS and FBED-CART-Sum show similar results as wkNN, in this section, we explain the performance difference between DepAD algorithms and wkNN. The following analysis is conducted with both FBED-CART-PS and FBED-CART-Sum, and the results are very similar. We only present the analysis based on FBED-CART-PS in the paper to save space.
	
	The 32 datasets are divided into two groups: low-dimensional and high-dimensional datasets. To achieve this, we used the median dimension of the datasets, which was 31. Datasets with 31 or fewer variables were categorized into the low-dimensional group, while datasets with more than 31 variables were placed in the high-dimensional group. As a result, both groups contained an equal number of datasets, ensuring the comparability of the $ p $-values. Subsequently, we conducted Wilcoxon signed-ranks tests on the experimental results (ROC AUC and AP) of the two groups, respectively. 
	
	As demonstrated in Table \ref{tbl:p_depad_wknn}, the $ p $-values of the high-dimensional group were consistently lower than the $ p $-values of the low-dimensional group, for both ROC AUC and AP. These findings suggest that FBED-CART-PS may possess advantages when applied to higher-dimensional datasets.
	
	\begin{table}[htbp] \centering
		\caption{p-value of Wilcoxon singed ranking test between FBED-CART-PS and wkNN.} 
		\label{tbl:p_depad_wknn}
		\resizebox{0.95\columnwidth}{!}{	
			\begin{tabular}{ c| ccc }
				\toprule
				\textbf{metric} & \textbf{all datasets} & \textbf{low-dimensional} & \textbf{high-dimensional} \\
				\midrule
				ROC AUC & 0.848 & 0.835 & 0.661 \\
				AP & 0.54 & 0.619 & 0.463 \\
				\bottomrule
			\end{tabular}
		}
	\end{table}
	
	\subsection{Impact of Data Characteristics on the Effectiveness of DepAD Algorithms} \label{sec:sensitivity}
	In this section, we explore the influence of various data characteristics on the performance of DepAD algorithms. We focus on data characteristics that are known to affect the effectiveness of anomaly detection, such as the dataset dimension, as well as some characteristics that hold particular significance for dependency-based algorithms, including the percentage of independent variables. Our analysis encompasses a total of six dataset characteristics.
	
	\begin{enumerate}
		\item \textit{Dimension}: the dimension (i.e., number of variables) of a dataset
		
		\item \textit{Correlation}: the average correlations of all pairs of variables
		
		\item \textit{Sparseness}: the sparseness of a dataset. Sparseness is a concept from graph theory. If we use a Bayesian network to represent the joint distribution of a dataset, the fewer edges the Bayesian network has, the sparser the dataset is. Hence, the average degree, i.e., the average size of the parent and child (PC) sets, of all variables in the Bayesian network learned from a dataset is utilized to represent the sparseness of the dataset. To be consistent with the semantics of sparseness, the value of the sparseness is set to the negative value of its average degree.
		
		\item \textit{IndepVar}: the percentage of independent variables. Independent variables refer to the variables that are not linked to any other variables in a Bayesian network. Therefore, IndepVar characteristic is calculated as the percentage of variables with no PC variables.
		
		\item The percentage of anomalies.
		
		\item The number of noisy variables: noisy variables are the variables that are unrelated to the data generation process. Research~\cite{chang2023data, han2022adbench, lazarevic2005feature} has shown that these variables can hide the characteristics of anomalies, making anomaly detection more challenging.
	\end{enumerate}
	
	We analyze the impacts of the first four characteristics on real-world datasets using the experimental results of the 32 datasets. For the last two characteristics, we conduct the analysis on synthetic datasets to ensure better control over these characteristics, as they are difficult to manipulate in real-world datasets.
	
	\subsubsection*{\textbf{Evaluation with Real-World Data.}} 
\begin{figure}[h] 
		\centering 
		\begin{subfigure}{0.45\columnwidth}
			\centering
			\includegraphics[width=1\columnwidth]{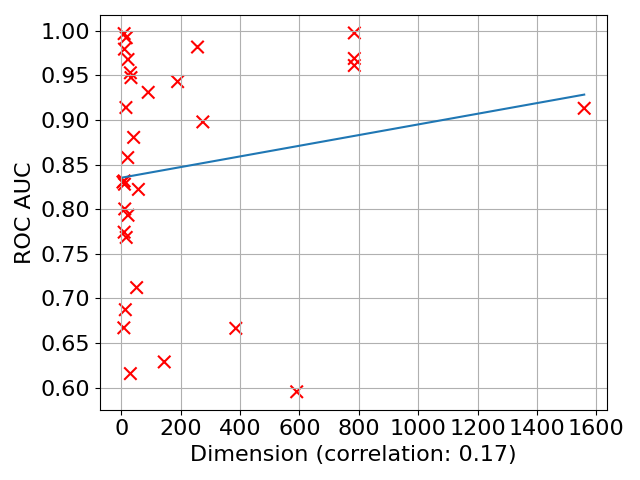}
			\caption{Dimension}
			\label{fig:dimension_roc}
		\end{subfigure}
		\begin{subfigure}{0.45\columnwidth}
			\centering
			\includegraphics[width=1\columnwidth]{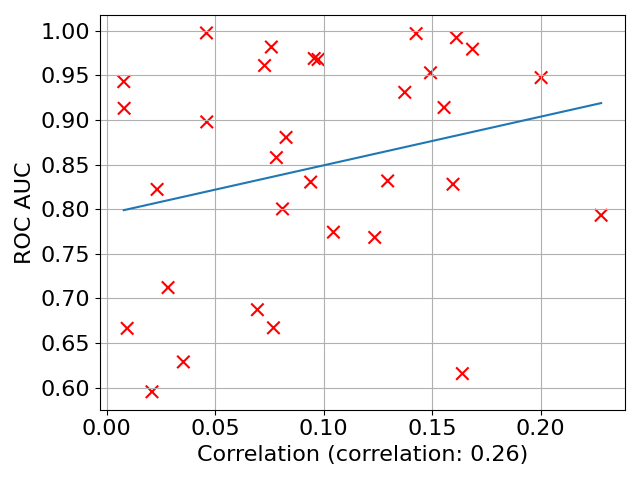}
			\caption{Correlation}
			\label{fig:correlation_roc}
		\end{subfigure}
		\begin{subfigure}{0.45\columnwidth}
			\centering
			\includegraphics[width=1\columnwidth]{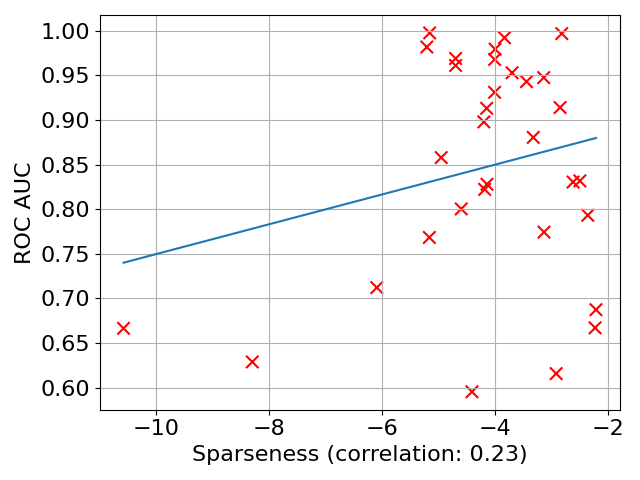}
			\caption{Sparseness}
			\label{fig:sparseness_roc}
		\end{subfigure}
		\begin{subfigure}{0.45\columnwidth}
			\centering
			\includegraphics[width=1\columnwidth]{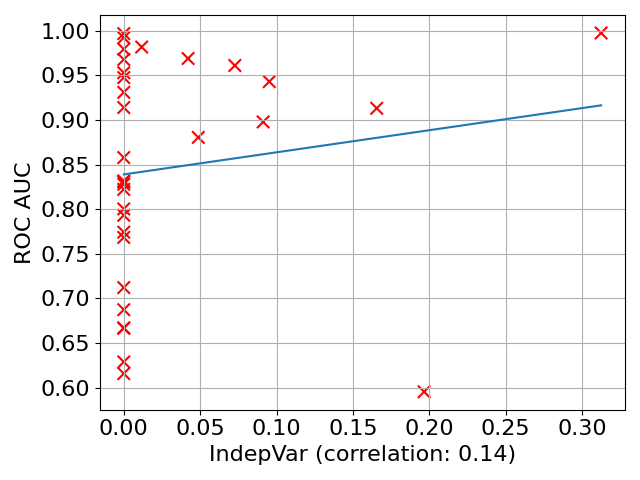}
			\caption{IndepVar}
			\label{fig:indepvar_roc}
		\end{subfigure}
		\caption{Impact of data characteristics on the performance of FBED-CART-PS (ROC AUC). Each subfigure displays 32 red crosses, each representing the result of a dataset. The x-coordinate represents the value of the characteristic named in sub-figure's title, and the y-coordinate indicates ROC AUC. A learned regression line is represented by the blue line. The Pearson correlation between ROC AUC and each characteristic is provided in the label of the x-axis for each subfigure.}
		\label{fig:char_imporantce_roc} 
	\end{figure}

\begin{figure}[h] 
		\centering 
		\begin{subfigure}{0.45\columnwidth}
			\centering
			\includegraphics[width=1\columnwidth]{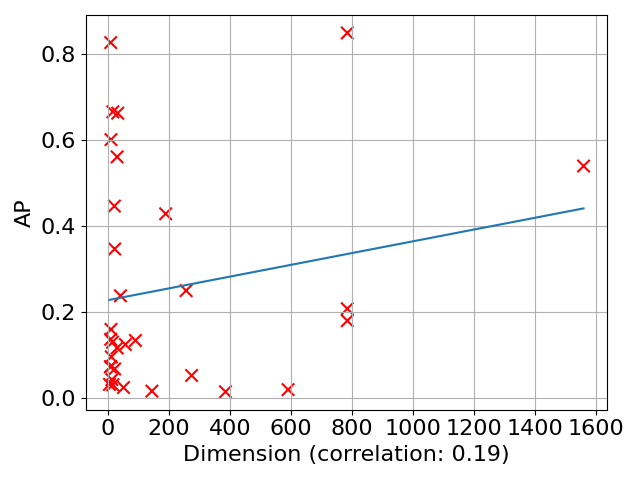}
			\caption{Dimension}
			\label{fig:dimension_ap}
		\end{subfigure}
		\begin{subfigure}{0.45\columnwidth}
			\centering
			\includegraphics[width=1\columnwidth]{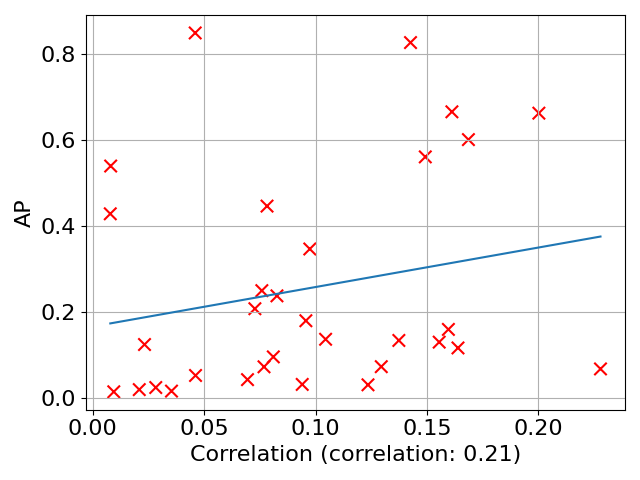}
			\caption{Correlation}
			\label{fig:correlation_ap}
		\end{subfigure}
		\begin{subfigure}{0.45\columnwidth}
			\centering
			\includegraphics[width=1\columnwidth]{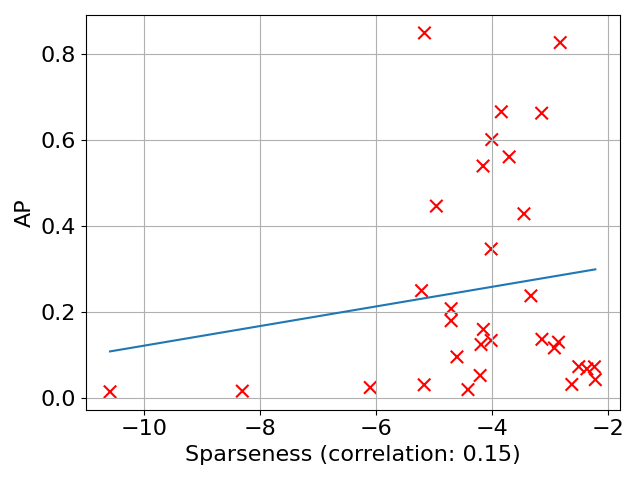}
			\caption{Sparseness}
			\label{fig:sparseness_ap}
		\end{subfigure}
		\begin{subfigure}{0.45\columnwidth}
			\centering
			\includegraphics[width=1\columnwidth]{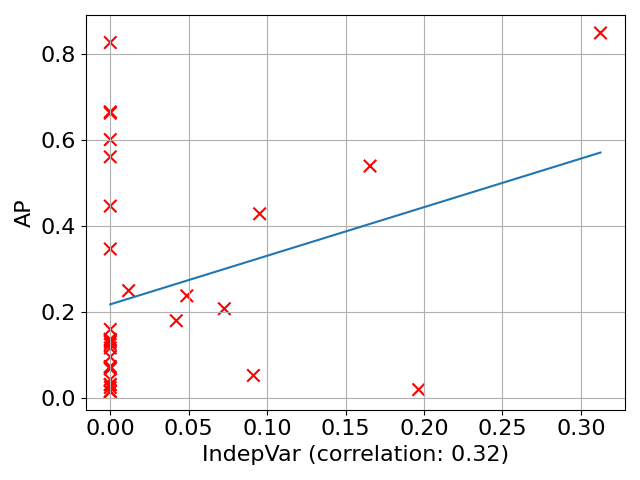}
			\caption{IndepVar}
			\label{fig:indepvar_ap}
		\end{subfigure}
		\caption{Importance of data characteristics to anomaly detection results (AP). Each subfigure displays 32 red crosses, each representing the result of a dataset. The x-coordinate represents the value of the characteristic named in the sub-figure's title, and the y-coordinate indicates AP. A learned regression line is represented by the blue line. The Pearson correlation between AP and each characteristic is provided in the label of the x-axis for each subfigure.}
		\label{fig:char_imporantce_ap} 
	\end{figure}

	The impact of the first four characteristics on anomaly detection results, regarding ROC AUC and AP, is depicted in Figures \ref{fig:char_imporantce_roc} and \ref{fig:char_imporantce_ap}, respectively. The experiments utilize the FBED-CART-PS anomaly detection method.
	
	In Figure \ref{fig:char_imporantce_roc}, the ROC AUCs of FBED-CART-PS exhibit very weak positive correlation with the four characteristics. The trend is similar for the impacts of these characteristics on AP. As depicted in Figure \ref{fig:char_imporantce_ap}, the APs also show weak positive correlations with the four characteristics. In conclusion, FBED-CART-PS demonstrates limited sensitivity to the dimensionality, correlation, sparseness, and percentage of independent variables within the dataset.
	
	DepAD algorithms, such as FBED-CART-PS, rely on identifying anomalies based on the dependency among variables. If an anomaly occurs solely on independent variables that have no dependency on other variables, it may not be detected by dependency-based algorithms. However, the analysis of the 32 real-world datasets does not substantiate this assumption. We speculate that the reason behind this discrepancy is the rarity of anomalies that solely affect independent variables in these datasets. It appears that such isolated anomalies are infrequent within the 32 real-world datasets, which might explain why the percentage of independent variables does not significantly impact the effectiveness of FBED-CART-PS.
	
	\subsubsection*{\textbf{Evaluation with Synthetic Data.}} 
	The synthetic datasets used in our experiments are generated using the ARTH150 benchmark Bayesian network from the Bnlearn repository~\cite{bnlearn}. Each dataset comprises 5000 objects and 107 variables, known as original variables, with Gaussian distributions whose means depend linearly on their parent variables in the Bayesian network. For each sensitivity experiment, we create 20 datasets, and the average ROC AUC and AP are reported, along with their 95\% confidence interval.
	
	To study the impact of the anomaly percentage, we randomly select a certain percentage of objects and 10\% of variables in each dataset to inject anomalous values. The percentage of anomalies range from 1\% to 10\%. The anomalies are generated by adding a perturbation to the original values and also ensure the perturbed values remaining within the observed range of the corresponding variables. Specifically, for a selected variable $X_j$, the perturbation, $d$ is calculated as $d=(max(X_j)-min(X_j))/2$, where $\min(X_j)$ and $max(X_j)$ represent the minimum and maximum observed values of $X_j$, respectively. Then, for an object $\mathbf{x}_{i*}$, its anomalous value on variable $X_j$ is computed as $(x_{ij} + d) \bmod max(X_j) + \lfloor \frac{min(X_j) * (x_{ij} + d)}{max(X_j)} \rfloor$, where $\bmod$ is the modulus operator.
	
\begin{figure}[h] 
		\centering 
		\begin{subfigure}{0.45\columnwidth}
			\centering
			\includegraphics[width=1\columnwidth]{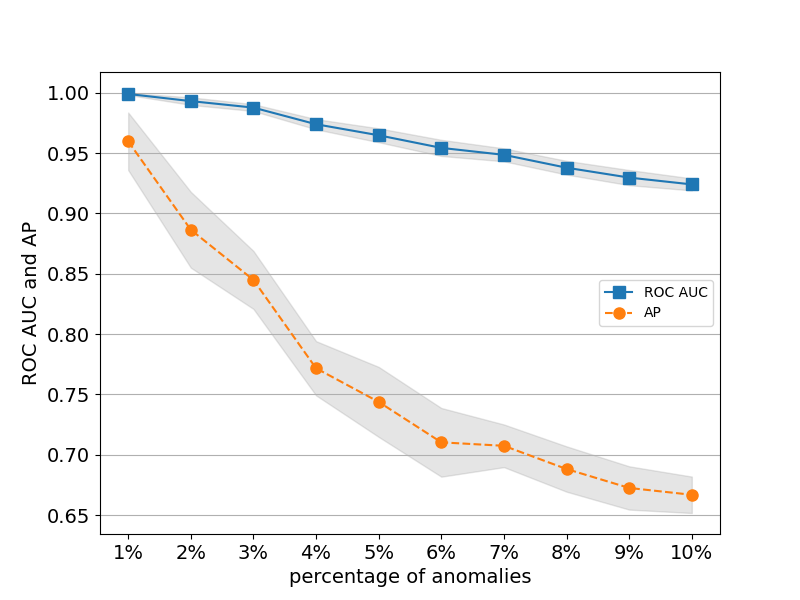}
			\caption{Percentage of Anomalies}
			\label{fig:sen_ratio_ano}
		\end{subfigure}
		\begin{subfigure}{0.45\columnwidth}
			\centering
			\includegraphics[width=1\columnwidth]{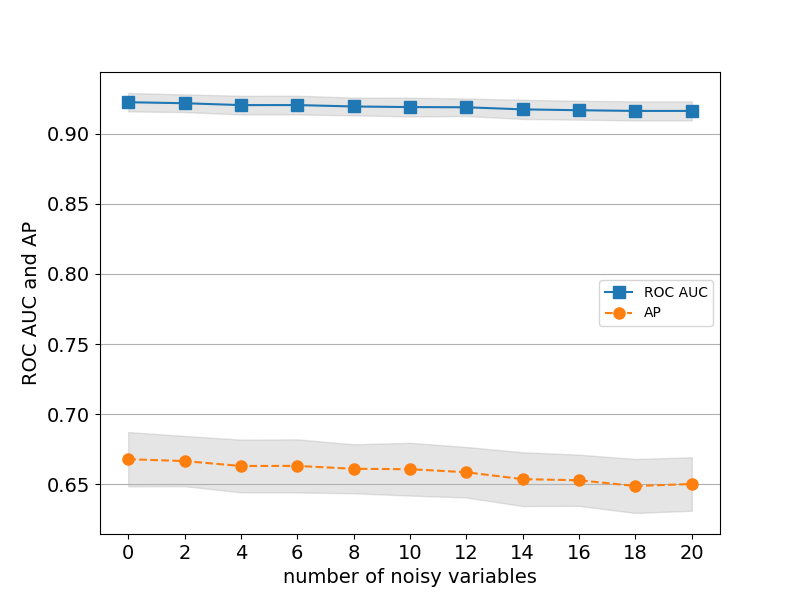}
			\caption{Number of Noisy Variables}
			\label{fig:sen_noisy_var}
		\end{subfigure}
		\caption{Sensitivity test results on synthetic data. Each subfigure displays two curves: the blue curve with solid squares represents the ROC AUC, while the orange dashed curve with dots represents the AP. The gray shaded area represents the 95\% confidence interval.}
		\label{fig:sen_syn} 
	\end{figure}

	Regarding the experiments on noisy variables, we introduce noisy variables into the synthetic datasets following the process in existing literature~\cite{lazarevic2005feature}. Specifically, to ensure minimal dependency between the noisy and the original variables, the values of these noisy variables are drawn from a uniform distribution between 0 and 1, and their correlation with the original variables is controlled below 0.1. The experimental results are presented in Figure \ref{fig:sen_syn}. 
	
	In Figure \ref{fig:sen_ratio_ano}, we observe that the ROC AUCs and APs of FBED-CART-PS consistently decrease as the percentage of anomalies increases. The narrower 95\% confidence interval of ROC AUC compared to AP suggests greater stability in the ROC AUC measurements. These results are expected since anomalies in the datasets can adversely affect the accuracy of relevant variable selection and prediction models, leading to a decline in anomaly detection performance.
	
	Figure \ref{fig:sen_noisy_var} demonstrates the impact of the number of noisy variables, ranging from 0 to 20, accounting for 0\% to 18\% of the original variables, with a fixed percentage of anomalies at 10\%. FBED-CART-PS exhibits low sensitivity to noisy variables in terms of both ROC AUC and AP. This behavior can be attributed to the relevant variable selection step within the DepAD framework, which effectively excludes noisy variables from the prediction models, ensuring they do not compromise the accuracy of expected value predictions.
	
	\subsection{Efficiency of DepAD Algorithms} \label{sec:exp_efficiency}
	Computational costs of DepAD algorithms mainly come from the relevant variable selection and prediction model training phases. Different techniques have different time complexities. We summarize the average running times (in seconds) of the five relevant variable selection techniques and the five prediction models. It is noted that for FBED and HITON-PC, we use the implementations in CausalFS~\cite{yu2019causality, yu2018unified} that is based on C language, while the implementations of the other relevant variables selection techniques use R language. Therefore, the running times of the relevant variable selection phase are not directly comparable, but we still present them here for the reference of the readers.
	
	\begin{table*}[htbp]
		\centering
		\renewcommand{\arraystretch}{0.8}
		\caption{Average running time (in seconds) of the relevant variable selection techniques and prediction models.} 
		\label{tbl:running_time}
		\resizebox{0.7\textwidth}{!}{
			\begin{threeparttable} 
				\begin{tabular}{ l | ccccc | rrrrr }
					\toprule 
					& \multicolumn{5}{c|}{ \textbf{Relevant Variable Selection} } & 
					\multicolumn{5}{c}{ \textbf{Prediction Model Training}} \\ [0.3ex] 
					\cmidrule(l){2-6} \cmidrule(l){7-11} 
					\raisebox{1.5ex}{\textbf{Dataset}}& \textbf{FBED} \tnote{a} & \textbf{HITON-PC}\tnote{a} & \textbf{IEPC} & \textbf{DC} & \textbf{MI} & \textbf{mCART} & \textbf{CART} & \textbf{Linear} & \textbf{Lasso} & \textbf{Ridge} \\
					\midrule 
					Wilt & 0 \tnote{*} & 0 \tnote{*} & 14 & 25 & 0.08 & 1.8 & 1.5 & 0.15 & 0.21 & 0.14 \\
					Pima & 0 \tnote{*} & 0 \tnote{*} & 2.3 & 0.19 & 0.06 & 1.2 & 0.76 & 0.14 & 0.14 & 0.14 \\
					WBC & 0 \tnote{*} & 0 \tnote{*} & 1.8 & 0.15 & 0.08 & 1 & 0.76 & 0.16 & 0.15 & 0.15 \\
					Stamps & 0 \tnote{*} & 0 \tnote{*} & 3.4 & 0.4 & 0.07 & 1.1 & 0.77 & 0.15 & 0.14 & 0.14 \\
					Glass & 0 \tnote{*} & 0 \tnote{*} & 1.6 & 0.15 & 0.07 & 1.1 & 0.7 & 0.14 & 0.14 & 0.14 \\
					Gamma & 0 \tnote{*} & 0 \tnote{*} & 162 & 472 & 0.51 & 9.4 & 10 & 0.19 & 0.17 & 0.17 \\
					PageBlocks& 0 \tnote{*} & 0 \tnote{*} & 33 & 4.4 & 0.24 & 4.1 & 3 & 0.16 & 0.16 & 0.16 \\
					Wine & 0 \tnote{*} & 0 \tnote{*} & 37 & 2.1 & 0.29 & 5.2 & 4.1 & 0.16 & 0.16 & 0.16 \\
					HeartDisease & 0 \tnote{*} & 0 \tnote{*} & 2.9 & 0.25 & 0.17 & 1.4 & 1 & 0.16 & 0.15 & 0.15 \\
					Leaf & 0 \tnote{*} & 0 \tnote{*} & 10 & 1.1 & 0.21 & 1.7 & 1.3 & 0.16 & 0.15 & 0.15 \\
					Letter & 0 \tnote{*} & 0 \tnote{*} & 11 & 0.75 & 0.26 & 2.6 & 1.6 & 0.16 & 0.15 & 0.15 \\
					PenDigits & 0 \tnote{*} & 0 \tnote{*} & 133 & 8.2 & 1.1 & 2.4 & 1.5 & 0.25 & 0.19 & 0.2 \\
					Waveform & 0 \tnote{*} & 0.02 & 62 & 6.6 & 0.88 & 5.3 & 6.1 & 0.19 & 0.18 & 0.18 \\
					Cardiotocography & 0 \tnote{*} & 0 \tnote{*} & 45 & 2.8 & 0.65 & 5 & 3.6 & 0.18 & 0.17 & 0.17 \\
					Parkinson & 0 \tnote{*} & 0 \tnote{*} & 11 & 1 & 0.44 & 2.3 & 1.7 & 0.19 & 0.17 & 0.18 \\
					BreastCancer & 0 \tnote{*} & 0 \tnote{*} & 47 & 5.5 & 0.84 & 5 & 3.2 & 0.22 & 0.21 & 0.19 \\
					WBPC & 0 \tnote{*} & 0 \tnote{*} & 23 & 2.2 & 0.79 & 3.3 & 2.7 & 0.21 & 0.19 & 0.18 \\
					Ionosphere & 0 \tnote{*} & 0 \tnote{*} & 34 & 3.5 & 0.91 & 3.8 & 2.7 & 0.2 & 0.18 & 0.17 \\
					Biodegradation & 0 \tnote{*} & 0 \tnote{*} & 45 & 3.2 & 1.4 & 4.8 & 3.2 & 0.21 & 0.19 & 0.18 \\
					Bank & 0 \tnote{*} & 0.16 & 582 & 30 & 6 & 3.3 & 2 & 0.38 & 0.26 & 0.29 \\
					Spambase & 0 \tnote{*} & 0.06 & 555 & 31 & 6.3 & 2.8 & 2.1 & 0.38 & 0.29 & 0.28 \\
					Libras & 0 \tnote{*} & 0 \tnote{*} & 316 & 30 & 7.4 & 2 & 1.3 & 0.49 & 0.27 & 0.31 \\
					Aid362 & 0.58 & 4.3 & 5647 & 417 & 48 & 9.7 & 5.8 & 2.3 & 2 & 1.4 \\
					Backdoor & 1.9 & 1.9 & 4610 & 758 & 38 & 72 & 87 & 113 & 89 & 78 \\
					CalTech16 & 0.51 & 0.77 & 87 & 8.2 & 2.7 & 5.4 & 2.4 & 1.3 & 1 & 0.82 \\
					Arrhythmia & 0.28 & 0.28 & 56 & 5.1 & 2.1 & 4.7 & 2.5 & 1.2 & 2.7 & 0.77 \\
					Census & 6.9 & 502 & 10638 & 709 & 117 & 332 & 150 & 272 & 94 & 187 \\
					Secom & 1.8 & 4.1 & 1007 & 289 & 15 & 20 & 22 & 20 & 37 & 13 \\
					MNIST & 6.3 & 4.7 & 530 & 50 & 16 & 17 & 8.1 & 10 & 25 & 5.5 \\
					calTech28 & 12 & 10 & 664 & 60 & 21 & 26 & 6.6 & 5.4 & 5.3 & 2.9 \\
					Fashion & 29 & 11 & 3759 & 272 & 82 & 32 & 47 & 33 & 12 & 14 \\
					Ads & 27 & 296 & 20090 & 1064 & 369 & 77 & 52 & 189 & 8.4 & 19 \\ 
					\midrule
					\textbf{Average} & \textbf{2.7} & \textbf{26} & \textbf{1538} & \textbf{133} 
					& \textbf{23} & \textbf{21} & \textbf{14} & \textbf{20} & \textbf{8.8} & \textbf{10} \\ 
					\bottomrule
				\end{tabular}
				\begin{tablenotes}\footnotesize
					\item[a] These two techniques are implemented using CausalFS~\cite{yu2019causality, yu2018unified} written in C, while others use R language.
					\item[*] If a running time is less than 0.01, it is displayed as 0 in the table.
				\end{tablenotes}
			\end{threeparttable} 
		}
	\end{table*}
	
	The running times on the 32 datasets and their average values are shown in Table \ref{tbl:running_time}. Comparing the five methods, FBED is the most efficient, with an average running time of 2.7 seconds, followed by MI at 23 seconds, HITON-PC at 26 seconds, DC at 133 seconds, and IEPC being the most time-consuming at 1538 seconds. Notably, IEPC shows the longest average running time, particularly for large-sized datasets like Ads, Census, and Backdoor. As for the five prediction models, their computational costs are moderate, ranging from 8.8 to 21 seconds on average.
	
	\begin{table*}[htbp]
		\centering
		\renewcommand{\arraystretch}{0.8}
		\caption{Average running time (in seconds) of the DepAD algorithms and benchmark methods.} 
		\label{tbl:rt_methods}
		\resizebox{0.8\textwidth}{!}{
			\begin{threeparttable} 
				\begin{tabular}{ l | cc | rrrrrrrrr }
					\toprule 
					& \multicolumn{2}{|c|}{ \textbf{DepAD}} &
					\multicolumn{9}{c}{ \textbf{Benchmark Methods}} \\
					\cmidrule(l){2-3}\cmidrule(l){4-12}
					\textbf{Dataset} 
					& \textbf{1} \tnote{*} & \textbf{2}\tnote{*} 
					& \textbf{LOF} & \textbf{wkNN} & \textbf{FastABOD} & \textbf{iForest} & \textbf{MBOM} & \textbf{SOD}
					& \textbf{OCSVM} & \textbf{ALSO} & \textbf{COMBN} \\
					\midrule 
					Wilt & 1.4 & 1.4 & 0.38 & 0.16 & 6.3 & 0.49 & 2.1 & 132 & 0.03 & 18 & 0.02 \\
					Pima & 0.63 & 0.64 & 0.04 & 0.02 & 0.31 & 0.37 & 0.42 & 4.7 & 0 & 4.6 & 0.01 \\
					WBC & 0.58 & 0.58 & 0.04 & 0.01 & 0.27 & 0.36 & 0.55 & 4.1 & 0 & 4.6 & 0.02 \\
					Stamps & 0.62 & 0.6 & 0.03 & 0.01 & 0.19 & 0.36 & 0.36 & 2.7 & 0 & 4.4 & 0.01 \\
					Glass & 0.61 & 0.62 & 0.02 & 0.01 & 0.12 & 0.36 & 0.25 & 1.7 & 0 & 3.2 & 0.01 \\
					Gamma & 9.7 & 9.7 & 1.3 & 0.68 & 36 & 1.4 & 12 & 852 & 0.3 & 178 & 0.3 \\
					PageBlocks& 3.4 & 3.3 & 0.41 & 0.18 & 7.2 & 0.58 & 4.5 & 157 & 0.06 & 53 & 0.17 \\
					Wine & 4.7 & 4.7 & 0.54 & 0.31 & 6.3 & 0.59 & 5.9 & 156 & 0.05 & 66 & 0.22 \\
					HeartDisease & 0.73 & 0.71 & 0.02 & 0.01 & 0.17 & 0.52 & 0.36 & 2.4 & 0 & 6 & 0.02 \\
					Leaf & 0.95 & 0.95 & 0.03 & 0.01 & 0.22 & 0.37 & 0.51 & 3.6 & 0 & 9.8 & 0.03 \\
					Letter & 1.7 & 1.7 & 0.07 & 0.04 & 0.53 & 0.38 & 1.2 & 10 & 0.01 & 19 & 0.15 \\
					PenDigits & 1.5 & 1.5 & 1 & 0.5 & 25 & 1.4 & 17 & 578 & 0.25 & 263 & 2.1 \\
					Waveform & 3.6 & 3.3 & 0.22 & 0.14 & 1.9 & 0.47 & 3 & 34 & 0.03 & 94 & 0.24 \\
					Cardiotocography & 3.5 & 3.5 & 0.18 & 0.09 & 1.6 & 0.63 & 3.5 & 32 & 0.02 & 53 & 0.43 \\
					Parkinson & 1.3 & 1.2 & 0.01 & 0.01 & 0.1 & 0.36 & 0.44 & 1.8 & 0 & 11 & 0.05 \\
					BreastCancer & 2.4 & 2.4 & 0.03 & 0.02 & 0.28 & 0.37 & 1.3 & 5.5 & 0.01 & 33 & 0.31 \\
					WBPC & 1.8 & 1.6 & 0.01 & 0.01 & 0.12 & 0.54 & 0.6 & 2.2 & 0 & 15 & 0.14 \\
					Ionosphere & 2.2 & 2.1 & 0.02 & 0.01 & 0.18 & 0.37 & 0.97 & 3.5 & 0 & 27 & 0.33 \\
					Biodegradation & 2.6 & 2.6 & 0.04 & 0.02 & 0.28 & 0.38 & 1.9 & 6.6 & 0.01 & 49 & 0.57 \\
					Bank & 1.2 & 1.3 & 0.75 & 0.51 & 7.1 & 1.1 & 19 & 161 & 0.19 & 546 & 7.7 \\
					Spambase & 1.2 & 1.2 & 0.39 & 0.25 & 4.7 & 0.75 & 14 & 100 & 0.17 & 615 & 3 \\
					Libras & 1.2 & 1.2 & 0.03 & 0.02 & 0.37 & 0.42 & 5.7 & 11 & 0.02 & 212 & 1.7 \\
					Aid362 & 5.3 & 5.5 & 1.8 & 1.6 & 9.6 & 2.1 & 138 & 292 & 1.6 & 4154 & 561 \\
					Backdoor & 42 & 43 & 13 & 11 & - \tnote{a}& 177 & 1077 & - \tnote{a}& 47 & - \tnote{a}& - \tnote{a} \\
					CalTech16 & 2.7 & 2.8 & 0.16 & 0.14 & 1.5 & 0.8 & 36 & 68 & 0.28 & - \tnote{a}& - \tnote{a} \\
					Arrhythmia & 2.2 & 2.3 & 0.05 & 0.04 & 0.63 & 0.53 & 16 & 27 & 0.08 & 1299 & 6.9 \\
					Census & 106 & 104 & 125 & 120 & - \tnote{a}& 360 & 2064 & - \tnote{a}& 1304 & - \tnote{a}& - \tnote{a} \\
					Secom & 8.1 & 8.2 & 0.83 & 0.75 & 6.1 & 2.3 & 91 & 226 & 3.8 & - \tnote{a}& - \tnote{a} \\
					MNIST & 12 & 12 & 0.37 & 0.37 & 3.9 & 1.7 & 188 & 175 & 1.6 & - \tnote{a}& - \tnote{a} \\
					calTech28 & 17 & 18 & 0.31 & 0.27 & 3.5 & 1.6 & 1570 & 181 & 1.9 & - \tnote{a}& - \tnote{a} \\
					Fashion & 38 & 38 & 0.44 & 0.43 & 5.3 & 2.2 & 732 & 234 & 1.8 & - \tnote{a}& - \tnote{a} \\
					Ads & 41 & 37 & 3.3 & 3.6 & 32 & 32 & 852 & 792 & 14 & - \tnote{a}& - \tnote{a} \\
					\midrule
					\textbf{Average} & \textbf{10.1} & \textbf{9.9} & \textbf{4.7} 
					& \textbf{4.4} & \textbf{5.4} & \textbf{18} & \textbf{214} & \textbf{142} 
					& \textbf{43} & \textbf{322} & \textbf{24}\\
					\bottomrule
				\end{tabular}
				\begin{tablenotes}
					\item[*] 1: FBED-CART-PS; $\qquad$ 2: FBED-CART-Sum	 
					\item[a] The running time is longer than 4 hours.
				\end{tablenotes}
			\end{threeparttable} 
		}
	\end{table*}
	
	The overall running time of the two DepAD algorithms and the nine benchmark methods are presented in Table \ref{tbl:rt_methods}. In general, the two DepAD algorithms have high efficiency. In the nine benchmark methods, FastABOD, ALSO, SOD and COMBN could not finish in four hours on some datasets.
	
	\subsection{Findings and Discussion}	
	The evaluation of the DepAD framework has provided valuable insights into its performance and effectiveness for dependency-based anomaly detection. The key findings are as follows:
	
	\begin{enumerate}	
		\item \textbf{Effectiveness}: The two DepAD algorithms, FBED-CART-PS, and FBED-CART-Sum, demonstrate superior performance over nine state-of-the-art anomaly detection methods in the majority of cases. The two DepAD methods do not outperform wkNN. However, they show advantages over wkNN in higher dimensional datasets in terms of both ROC AUC and AP.
		
		\item \textbf{Efficiency}: The two DepAD algorithms generally demonstrate high efficiency compared to several benchmark methods.
		
		\item \textbf{Feature Selection Techniques}: Causal feature selection techniques, such as FBED and HITON-PC, generally lead to better anomaly detection performance within the DepAD framework.
		
		\item \textbf{Prediction Models}: Tree-based models, such as CART and mCART, outperform linear models due to their ability to handle complex relationships. In contrast, Lasso yields the worst results in among all instantiations.
		
		\item \textbf{Scoring Techniques}: Summation-based techniques, such as RZPS, PS and Sum, produce the better results than GS and Max.
		
		\item \textbf{Sensitivity Experiments}: DepAD algorithms are not sensitive to the average correlation, sparseness, or dimensionality of datasets. DepAD methods exhibit stability when data contains noisy variables. However, the percentage of anomalies can negatively affect their effectiveness.		
	\end{enumerate}
	
	A limitation of the DepAD algorithms is their inability to detect or interpret anomalies that only affect independent variables. However, this situation is relatively rare in the evaluated real-world datasets. A potential improvement to address this limitation is adopting ensemble methods from both dependency-based and proximity-based perspectives, as suggested by prior studies~\cite{lu2018effective, aggarwal2017outlier, xie2021logdp}.
	
	\section{Interpretability of D\lowercase{ep}AD Algorithms} \label{sec:interpretation}
	The DepAD framework offers a significant advantage in providing meaningful explanations for identified anomalies, which plays a crucial role in understanding both the reported anomalies and the underlying data. In this section, we outline the process of interpreting an anomaly detected by a DepAD algorithm and demonstrate this through an illustrative example.
	
	To interpret an anomaly detected by DepAD, we begin by identifying variables with substantial dependency deviations. This is achieved by comparing the observed values of variables with their corresponding expected values. A larger deviation indicates a higher contribution of that variable to the anomaly. Furthermore, we gain insights into how the anomaly differs from normal behaviors by contrasting the observed dependency pattern with the normal dependency pattern between a variable and its relevant variables. The normal dependency pattern is represented by the expected value of a variable given the values of its relevant variables, while the observed value of the variable along with the values of its relevant variables constitutes the observed pattern. This comparison facilitates a comprehensive understanding of the anomaly's characteristics and the factors contributing to its detection.
	
\begin{figure}[htbp]
		\centering
		\includegraphics[width=0.9\columnwidth]{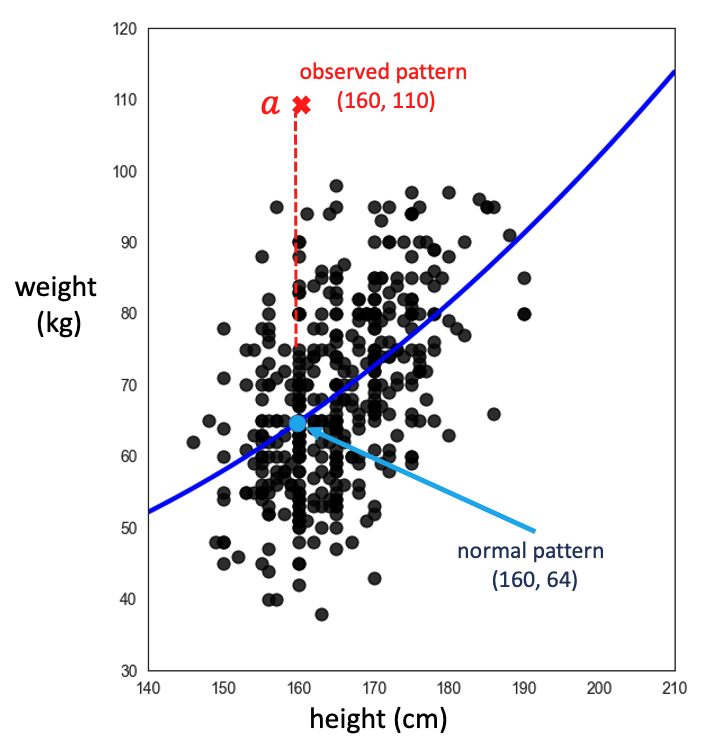}
		\caption{Illustration of interpretation of the DepAD algorithms.}
		\label{fig:interpret}
	\end{figure}

	For example, as shown in Figure \ref{fig:interpret}, in the dataset used in Example \ref{emp:obesity}, a person $ a $ has been identified as an anomaly by a DepAD method. From the dependency learned by the DepAD method, given a height of 160cm, the expected weight is 64kg. The normal pattern here is \textit{height=160cm $ \rightarrow $ weight=64kg}. Person $ a $ has a height of 160cm and a weight of 110kg. The observed pattern is \textit{height=160cm $\rightarrow$ weight=110kg}. The dependency deviation here is 110 - 64 = 56 kg. Then, the normal pattern, observed pattern and the dependency deviation can be used to explain why a is identified as an anomaly. In this case, we can see that $ a $ is flagged as an anomaly because he has a much larger weight than most people with the same height.
	
	In the following, we use a real-world dataset, the Zoo dataset, from the UCI machine learning repository~\cite{UciData} to illustrate how to interpret the anomalies found. The Zoo dataset contains information about 101 animals, each of which is described by 16 features. After applying the DepAD algorithm FBED-CART-PS to the dataset, it reports scorpion, platypus and sea snake as the top-3 anomalies. To interpret each of the three anomalies, we find and use its top-3 variables that contribute to the anomalousness most, i.e., the three variables with the highest derivation values. Then for each of the three variables of an anomaly, we pick its most relevant variable to examine the details of dependency violation. 
	
	The interpretations of the top-3 anomalies identified by FBED-CART-PS are presented in Table \ref{tbl:inter_Zoo}. For scorpion, the three variables backbone, eggs and milk contribute most to the anomalousness. For variable backbone, 73\% of the animals in the dataset follow the normal dependency; that is, if an animal has a tail, it would have a backbone. Scorpion has a tail but no backbone, and it is the only one with this dependency pattern violation (i.e., the observed pattern is not consistent with the expected pattern) in the dataset. In relation to the other two most contributing variables, eggs and milk, the normal dependency that is held by 57\% of animals in the dataset says if an animal does not produce milk, it lays eggs. Scorpion neither lays eggs nor produces milk, and only 2\% of the animals have this pattern. 
	
	\begin{table*}[htbp] \centering
		\renewcommand{\arraystretch}{0.8}
		\caption{Interpretations of top-$3$ anomalies detected by FBED-CART-PS in the Zoo dataset.} 
		\label{tbl:inter_Zoo}
		\resizebox{0.8\textwidth}{!}{
			\begin{threeparttable} 
				\begin{tabular}{ l lccc rr }
					\toprule
					\multirow{2}{*}{\textbf{anomaly}} 
					& \multicolumn{4}{c}{ \textbf{top-$3$ variables with most deviations}} 
					& \multicolumn{2}{c}{ \textbf{dependency violation}} \\
					\cmidrule(l){2-5}\cmidrule(l){6-7}
					
					& \textbf{variable} & $\boldsymbol{\delta}_j$ \tnote{a} & $\textbf{x}_j$ \tnote{b} & $\overline{\textbf{x}}_j$ \tnote{c} & \textbf{normal dependency pattern(\%)} \tnote{d} & \textbf{observed dependency pattern(\%)} \tnote{d} \\ 
					\midrule 
					
					\multirow{3}{*}{scorpion} & backbone & 9.2 & 0 & 1 
					& tail=1 $\rightarrow$ backbone=1 (73\%)
					& tail=1 $\rightarrow$ backbone=0 (0.9\%) \\ 
					
					& eggs & 5.3 & 0 & 0.9 
					& milk=0 $\rightarrow$ eggs=1 (57\%)
					& milk=0 $\rightarrow$ eggs=0 (2\%) \\
					
					& milk & 5.2 & 0 & 0.7
					& eggs=0 $\rightarrow$ milk=1 (57\%)
					& eggs=0 $\rightarrow$ milk=0 (2\%) \\
					\midrule
					
					\multirow{3}{*}{platypus} & toothed & 6.9 & 0 & 0.9 
					& milk=1 $\rightarrow$ toothed=1 (40\%) 
					& milk=1 $\rightarrow$ toothed=0 (1\%) \\ 
					
					& eggs & 6.1 & 1 & 0 
					& milk=1 $\rightarrow$ eggs=0 (40\%) & milk=1 	$\rightarrow$ eggs=1 (1\%) \\
					& milk & 5.5 & 1 & 0.2
					& eggs=1 $\rightarrow$ milk=0 (57\%)
					& eggs=1 $\rightarrow$ milk=1 (1\%) \\ 
					\midrule
					
					\multirow{3}{*}{seasnake} & eggs & 5.3 & 0 & 0.9 
					& milk=0 $\rightarrow$ eggs=1 (57\%) 
					& milk=0 $\rightarrow$ eggs=0 (2\%) \\ 
					& milk & 5.1 & 0 & 0.7 
					& eggs=0 $\rightarrow$ milk=1 (57\%) 
					& eggs=0 $\rightarrow$ milk=0 (2\%) \\
					& fins & 4.2 & 0 & 0.8
					& legs!=0 $\rightarrow$ fins=0 (76\%) \tnote{e}
					& legs=0 $\rightarrow$ fins=0 (7\%) \\ 
					\bottomrule
				\end{tabular}
				\begin{tablenotes}
					\item[a] The column shows the normalized deviations.
					\item[b] The column shows the observed values.
					\item[c] The column shows the expected values.
					\item[d] The \% after each pattern indicates the percentage of animals having the pattern in the Zoo dataset.
					\item[e] != means not equal to.
				\end{tablenotes}
			\end{threeparttable} 
		}
	\end{table*} 
	
	Platypus mainly violates the following normal dependencies: 1) if an animal produces milk, it likely has teeth; 2) if an animal produces milk, it is unlikely to lay eggs; and 3) if an animal lays eggs, it is unlikely to produce milk. Each of the three normal patterns is held by over 40\% of animals in the dataset. Platypus is the only animal that violates these dependencies (accounting for 1\%).
	
	The first normal dependency that the sea snake violates is the relationship between eggs and milk, and it is one of the two animals in the dataset that neither lay eggs nor produce milk. Another normal dependency is that if an animal has legs, it is unlikely to have fins (held by 76\% of the animals in the dataset), but sea snake neither has legs nor fins.
	
	\section{Conclusion} \label{sec:conclusion}
	In this paper, we introduce DepAD, a versatile framework for dependency-based anomaly detection. DepAD offers a general approach to construct effective, scalable, and flexible anomaly detection algorithms by leveraging off-the-shelf feature selection techniques and supervised prediction models for various data types and applications. Through systematic investigations, we explore the suitability of different techniques within the DepAD framework. Comprehensive experiments are conducted to empirically evaluate the performance of these techniques in individual phases as well as their combinations across multiple phases of DepAD.
	
	We compare two high-performing instantiations of DepAD, FBED-CART-PS and FBED-CART-Sum, against nine state-of-the-art anomaly detection methods across 32 commonly used datasets. The results demonstrate that DepAD algorithms consistently outperform existing methods in most cases. Moreover, the DepAD framework's high interpretability is highlighted through a case study.
	
	In conclusion, DepAD empowers the creation of dependency-based anomaly detection methods tailored to specific domain requirements. While this paper investigates several common techniques, the choices of off-the-shelf techniques need not be confined to the evaluated options. Incorporating domain knowledge during instantiation and technique selection can lead to optimal anomaly detection performance. The flexibility of DepAD allows for customization of anomaly detection solutions to suit particular domain characteristics and constraints effectively.
	
	\section*{Acknowledgments}
	

	\bibliographystyle{elsarticle-num} 
	\bibliography{reference}

\begin{thebibliography}{10}
\expandafter\ifx\csname url\endcsname\relax
  \def\url#1{\texttt{#1}}\fi
\expandafter\ifx\csname urlprefix\endcsname\relax\def\urlprefix{URL }\fi
\expandafter\ifx\csname href\endcsname\relax
  \def\href#1#2{#2} \def\path#1{#1}\fi

\bibitem{thudumu2020comprehensive}
S.~Thudumu, P.~Branch, J.~Jin, J.~Singh, A comprehensive survey of anomaly
  detection techniques for high dimensional big data, {Journal} of {Big} {Data}
  7 (2020) 1--30.

\bibitem{aggarwal2016outlier}
C.~C. Aggarwal, Outlier analysis, 2nd Edition, Springer Publishing Company,
  Incorporated, 2016.

\bibitem{chandola2009anomaly}
V.~Chandola, A.~Banerjee, V.~Kumar, Anomaly detection: A survey, {ACM}
  {Computing} {Surveys} ({CSUR}) 41~(3) (2009) 15.

\bibitem{lu2019LoPAD}
S.~Lu, L.~Liu, J.~Li, T.~D. Le, J.~Liu, Lopad: a local prediction approach to
  anomaly detection, in: {Proceedings} of {Pacific}-{Asia} {Conference} on
  {Knowledge} {Discovery} and {Data} {Mining}, Springer, 2020, pp. 660--673.

\bibitem{paulheim2015decomposition}
H.~Paulheim, R.~Meusel, A decomposition of the outlier detection problem into a
  set of supervised learning problems, {Machine} {Learning} 100~(2-3) (2015)
  509--531.

\bibitem{UciData}
G.~C. Dua~Dheeru, \href{http://archive.ics.uci.edu/ml}{{UCI} machine learning
  repository} (2017).
\newline\urlprefix\url{http://archive.ics.uci.edu/ml}

\bibitem{xie2021logdp}
Y.~Xie, H.~Zhang, B.~Zhang, M.~A. Babar, S.~Lu, Logdp: combining dependency and
  proximity for log-based anomaly detection, in: {Proceedings} of
  {International} {Conference} on {Service}-{Oriented} {Computing}, Springer,
  2021, pp. 708--716.

\bibitem{yan2022structural}
W.~Yan, D.~Chronopoulos, K.~Yuen, Y.~Zhu, Structural anomaly detection based on
  probabilistic distance measures of transmissibility function and statistical
  threshold selection scheme, {Mechanical} {Systems} and {Signal} {Processing}
  162 (2022) 108009.

\bibitem{sarmadi2020novel}
H.~Sarmadi, A.~Karamodin, A novel anomaly detection method based on adaptive
  mahalanobis-squared distance and one-class knn rule for structural health
  monitoring under environmental effects, {Mechanical} {Systems} and {Signal}
  {Processing} 140 (2020) 106495.

\bibitem{ramaswamy2000efficient}
S.~Ramaswamy, R.~Rastogi, K.~Shim, Efficient algorithms for mining outliers
  from large data sets, in: {Proceedings} of {ACM} {SIGMOD} {International}
  {Conference} on {Management} of {Data}, 2000, pp. 427--438.

\bibitem{breunig2000lof}
M.~M. Breunig, H.~Kriegel, R.~T. Ng, J.~Sander, Lof: identifying density-based
  local outliers, in: {Proceedings} of {ACM} {SIGMOD} {International}
  {Conference} on {Management} of {Data}, 2000, pp. 93--104.

\bibitem{angiulli2005outlier}
F.~Angiulli, C.~Pizzuti, Outlier mining in large high-dimensional data sets,
  {IEEE} {Transactions} on {Knowledge} and {Data} {Engineering} 17~(2) (2005)
  203--215.

\bibitem{tang2002enhancing}
J.~Tang, Z.~Chen, A.~W. Fu, D.~W. Cheung, Enhancing effectiveness of outlier
  detections for low density patterns, in: {Proceedings} of {Pacific}-{Asia}
  {Conference} on {Knowledge} {Discovery} and {Data} {Mining}, Springer, 2002,
  pp. 535--548.

\bibitem{zhang2009new}
K.~Zhang, M.~Hutter, H.~Jin, A new local distance-based outlier detection
  approach for scattered real-world data, in: {Proceedings} of {{P}}acific-Asia
  {C}onference on {K}nowledge {D}iscovery and {D}ata {M}ining, Springer, 2009,
  pp. 813--822.

\bibitem{kriegel2009loop}
H.~Kriegel, P.~Kr{\"o}ger, E.~Schubert, A.~Zimek, Loop: local outlier
  probabilities, in: {Proceedings} of the {18th} {ACM} {Conference} on
  {Information} and {Knowledge} {Management}, 2009, pp. 1649--1652.

\bibitem{huang2003cross}
Y.~Huang, W.~Fan, W.~Lee, P.~S. Yu, Cross-feature analysis for detecting ad-hoc
  routing anomalies, in: {Proceedings} of the {23rd} {International}
  {Conference} on {Distributed} {Computing} {Systems}, IEEE, 2003, pp.
  478--487.

\bibitem{noto2010anomaly}
K.~Noto, C.~Brodley, D.~Slonim, Anomaly detection using an ensemble of feature
  models, in: {Proceedings} of {IEEE} {International} {Conference} on {Data}
  {Mining}, IEEE, 2010, pp. 953--958.

\bibitem{noto2012frac}
K.~Noto, C.~Brodley, D.~Slonim, Frac: a feature-modeling approach for
  semi-supervised and unsupervised anomaly detection, {Data} {Mining} and
  {Knowledge} {Discovery} 25~(1) (2012) 109--133.

\bibitem{babbar2012mining}
S.~Babbar, S.~Chawla, Mining causal outliers using gaussian bayesian networks,
  in: {Proceedings} of {IEEE} {24th} {International} {Conference} on {Tools}
  with {Artificial} {Intelligence}, Vol.~1, IEEE, 2012, pp. 97--104.

\bibitem{liang2022advances}
W.~Liang, G.~A. Tadesse, D.~Ho, L.~Fei-Fei, M.~Zaharia, C.~Zhang, J.~Zou,
  Advances, challenges and opportunities in creating data for trustworthy ai,
  {Nature} {Machine} {Intelligence} 4~(8) (2022) 669--677.

\bibitem{huynh2024dagnosis}
N.~Huynh, J.~Berrevoets, N.~Seedat, J.~Crabb{\'e}, Z.~Qian, M.~van~der Schaar,
  Dagnosis: Localized identification of data inconsistencies using structures,
  in: {Proceedings} of the {27th} {International} {Conference} on {Artificial}
  {Intelligence} and {Statistics} {(AISTATS)}, Vol. 238, 2024.

\bibitem{liu2020energy}
W.~Liu, X.~Wang, J.~Owens, Y.~Li, Energy-based out-of-distribution detection,
  {Advances} in {Neural} {Information} {Processing} {Systems} 33 (2020)
  21464--21475.

\bibitem{zimek2012survey}
A.~Zimek, E.~Schubert, H.~Kriegel, A survey on unsupervised outlier detection
  in high-dimensional numerical data, {Statistical} {Analysis} and {Data}
  {Mining}: the {ASA} {Data} {Science} {Journal} 5~(5) (2012) 363--387.

\bibitem{yu2018markov}
K.~Yu, H.~Chen, Markov boundary-based outlier mining, {IEEE} {Transactions} on
  {Neural} {Networks} and {Learning} {Systems} 30~(4) (2018) 1259--1264.

\bibitem{kriegel2012outlier}
H.~Kriegel, P.~Kr{\"o}ger, E.~Schubert, A.~Zimek, Outlier detection in
  arbitrarily oriented subspaces, in: {Proceedings} of {IEEE} {12th}
  {International} {Conference} on {Data} {Mining}, IEEE, 2012, pp. 379--388.

\bibitem{kriegel2009outlier}
H.~Kriegel, P.~Kr{\"o}ger, E.~Schubert, A.~Zimek, Outlier detection in
  axis-parallel subspaces of high dimensional data, in: {Proceedings} of
  {Pacific}-{Asia} {Conference} on {Knowledge} {Discovery} and {Data} {Mining},
  Springer, 2009, pp. 831--838.

\bibitem{liu2008isolation}
F.~T. Liu, K.~M. Ting, Z.~Zhou, Isolation forest, in: {Proceedings} of the
  {Eighth} {IEEE} {International} {Conference} on {Data} {Mining}
  ({ICDM}'{08}), IEEE, 2008, pp. 413--422.

\bibitem{lazarevic2005feature}
A.~Lazarevic, V.~Kumar, Feature bagging for outlier detection, in:
  {Proceedings} of the {Eleventh} {ACM} {SIGKDD} {International} {Conference}
  on {Knowledge} {Discovery} in {Data} {Mining}, 2005, pp. 157--166.

\bibitem{yu2018unified}
K.~Yu, L.~Liu, J.~Li, A unified view of causal and non-causal feature
  selection, {ACM} {Transactions} on {Knowledge} {Discovery} {From} {Data}
  ({TKDD}) 15~(4) (2021) 1--46.

\bibitem{li2017feature}
J.~Li, K.~Cheng, S.~Wang, F.~Morstatter, R.~P. Trevino, J.~Tang, H.~Liu,
  Feature selection: A data perspective, {ACM} {Computing} {Surveys} ({CSUR})
  50~(6) (2017) 1--45.

\bibitem{yu2019causality}
K.~Yu, X.~Guo, L.~Liu, J.~Li, H.~Wang, Z.~Ling, X.~Wu, Causality-based feature
  selection: methods and evaluations, {ACM} {Computing} {Surveys} ({CSUR}),
  53~(5) (2020).

\bibitem{guyon2007causal}
I.~Guyon, C.~Aliferis, Causal feature selection, in: {Computational} {Methods}
  of {Feature} {Selection}, Chapman and Hall/CRC, 2007, pp. 79--102.

\bibitem{borboudakis2019forward}
G.~Borboudakis, I.~Tsamardinos, Forward-backward selection with early dropping,
  The {Journal} of {Machine} {Learning} {Research} 20~(1) (2019) 276--314.

\bibitem{aliferis2003hiton}
C.~F. Aliferis, I.~Tsamardinos, A.~Statnikov, Hiton: a novel markov blanket
  algorithm for optimal variable selection, in: {Amia} {Annual} {Symposium}
  {Proceedings}, Vol. 2003, American Medical Informatics Association, 2003,
  p.~21.

\bibitem{aragam2015concave}
B.~Aragam, Q.~Zhou, Concave penalized estimation of sparse gaussian bayesian
  networks., {Journal} of {Machine} {Learning} {Research} 16~(1) (2015)
  2273--2328.

\bibitem{yaramakala2005speculative}
S.~Yaramakala, D.~Margaritis, Speculative markov blanket discovery for optimal
  feature selection, in: {Proceedings} of the {5th} {International}
  {Conference} on {Data} {Mining} ({ICDM}), 2005, pp. 4--pp.

\bibitem{margaritis2000bayesian}
D.~Margaritis, S.~Thrun, Bayesian network induction via local neighborhoods,
  in: {Advances} in {Neural} {Information} {Processing} {Systems}, 2000, pp.
  505--511.

\bibitem{pena2005scalable}
J.~M. Pe{\~n}a, J.~Bj{\"o}rkegren, J.~Tegn{\'e}r, Scalable, efficient and
  correct learning of markov boundaries under the faithfulness assumption, in:
  {Proceedings} of {European} {Conference} on {Symbolic} and {Quantitative}
  {Approaches} {To} {Reasoning} and {Uncertainty}, Springer, 2005, pp.
  136--147.

\bibitem{battiti1994using}
R.~Battiti, Using mutual information for selecting features in supervised
  neural net learning, {IEEE} {Transactions} on {Neural} {Networks} 5~(4)
  (1994) 537--550.

\bibitem{estevez2009normalized}
P.~A. Est{\'e}vez, M.~Tesmer, C.~A. Perez, J.~M. Zurada, Normalized mutual
  information feature selection, {IEEE} {Transactions} on {Neural} {Networks}
  20~(2) (2009) 189--201.

\bibitem{peng2005feature}
H.~Peng, F.~Long, C.~Ding, Feature selection based on mutual information
  criteria of max-dependency, max-relevance, and min-redundancy, {IEEE}
  {Transactions} on {Pattern} {Analysis} and {Machine} {Intelligence} 27~(8)
  (2005) 1226--1238.

\bibitem{guo2008gait}
B.~Guo, M.~S. Nixon, Gait feature subset selection by mutual information,
  {IEEE} {Transactions} on {Systems}, {Man}, And {Cybernetics}-{Part} {A}:
  {Systems} and {Humans} 39~(1) (2008) 36--46.

\bibitem{nguyen2014effective}
X.~V. Nguyen, J.~Chan, S.~Romano, J.~Bailey, Effective global approaches for
  mutual information based feature selection, in: {Proceedings} of the {20th}
  {ACM} {SIGKDD} {International} {Conference} on {Knowledge} {Discovery} and
  {Data} {Mining}, 2014, pp. 512--521.

\bibitem{dash2003consistency}
M.~Dash, H.~Liu, Consistency-based search in feature selection, {Artificial}
  {Intelligence} 151~(1-2) (2003) 155--176.

\bibitem{arauzo2008consistency}
A.~Arauzo-Azofra, J.~M. Benitez, J.~L. Castro, Consistency measures for feature
  selection, {Journal} of {Intelligent} {Information} {Systems} 30~(3) (2008)
  273--292.

\bibitem{dodge2008coefficient}
Y.~Dodge, Coefficient of determination, The {Concise} {Encyclopedia} of
  {Statistics} (2008) 88--91.

\bibitem{song2012feature}
L.~Song, A.~Smola, A.~Gretton, J.~Bedo, K.~Borgwardt, Feature selection via
  dependence maximization., {Journal} of {Machine} {Learning} {Research} 13~(5)
  (2012).

\bibitem{kira1992practical}
K.~Kira, L.~A. Rendell, A practical approach to feature selection, in:
  {Machine} {Learning} {Proceedings} {1992}, Elsevier, 1992, pp. 249--256.

\bibitem{kononenko1994estimating}
I.~Kononenko, Estimating attributes: Analysis and extensions of relief, in:
  {Proceedings} of {European} {Conference} on {Machine} {Learning}, Springer,
  1994, pp. 171--182.

\bibitem{robnik1997adaptation}
M.~Robnik-{\v{S}}ikonja, I.~Kononenko, An adaptation of relief for attribute
  estimation in regression, in: {Proceedings} of the {Fourteenth}
  {International} {Conference} of {Machine} {Learning} ({ICML}), Vol.~5, 1997,
  pp. 296--304.

\bibitem{roy2012robustness}
M.~Roy, D.~Larocque, Robustness of random forests for regression, {Journal} of
  {Nonparametric} {Statistics} 24~(4) (2012) 993--1006.

\bibitem{breiman1984classification}
L.~Breiman, J.~Friedman, C.~J. Stone, R.~A. Olshen, Classification and
  regression trees, CRC press, 1984.

\bibitem{quinlan2014c4}
J.~R. Quinlan, C4. 5: programs for machine learning, Elsevier, 2014.

\bibitem{wang1996induction}
Y.~Wang, I.~H. Witten, Induction of model trees for predicting continuous
  classes, {Computer} {Science} {Working} {Papers} (1996).

\bibitem{kass1980exploratory}
G.~V. Kass, An exploratory technique for investigating large quantities of
  categorical data, {Journal} of the {Royal} {Statistical} {Society}: {Series}
  {C} ({Applied} {Statistics}) 29~(2) (1980) 119--127.

\bibitem{cohen1995fast}
W.~W. Cohen, Fast effective rule induction, in: {Proceedings} of {Machine}
  {Learning}, Elsevier, 1995, pp. 115--123.

\bibitem{michalski1986multi}
R.~S. Michalski, I.~Mozetic, J.~Hong, N.~Lavrac, The multi-purpose incremental
  learning system aq15 and its testing application to three medical domains,
  in: {Proceedings} of the {Association} {for} the {Advancement} of
  {Artificial} {Intelligence} ({Aaai}), Vol. 1986, 1986, pp. 1--041.

\bibitem{smyth1992information}
P.~Smyth, R.~M. Goodman, An information theoretic approach to rule induction
  from databases, {IEEE} {Transactions} on {Knowledge} and {Data} {Engineering}
  4~(4) (1992) 301--316.

\bibitem{langley1992analysis}
P.~Langley, W.~Iba, K.~Thompson, An analysis of bayesian classifiers, in:
  {Proceedings} of the {Association} {for} the {Advancement} of {Artificial}
  {Intelligence} ({Aaai}), Vol.~90, 1992, pp. 223--228.

\bibitem{ramoni2001robust}
M.~Ramoni, P.~Sebastiani, Robust bayes classifiers, {Artificial} {Intelligence}
  125~(1-2) (2001) 209--226.

\bibitem{heckerman1997bayesian}
D.~Heckerman, Bayesian networks for data mining, {Data} {Mining} and
  {Knowledge} {Discovery} 1~(1) (1997) 79--119.

\bibitem{seber2009multivariate}
G.~A. Seber, Multivariate observations, Vol. 252, John Wiley \& Sons, 2009.

\bibitem{hoerl1970ridge}
A.~E. Hoerl, R.~W. Kennard, Ridge regression: Biased estimation for
  nonorthogonal problems, {Technometrics} 12~(1) (1970) 55--67.

\bibitem{zou2005regularization}
H.~Zou, T.~Hastie, Regularization and variable selection via the elastic net,
  {Journal} of the {Royal} {Statistical} {Society}: {Series} {B} ({Statistical}
  {Methodology}) 67~(2) (2005) 301--320.

\bibitem{tibshirani1996regression}
R.~Tibshirani, Regression shrinkage and selection via the lasso, {Journal} of
  the {Royal} {Statistical} {Society}: {Series} {B} ({Methodological}) 58~(1)
  (1996) 267--288.

\bibitem{friedman1991multivariate}
J.~H. Friedman, Multivariate adaptive regression splines, The {Annals} of
  {Statistics} (1991) 1--67.

\bibitem{boser1992training}
B.~E. Boser, I.~M. Guyon, V.~N. Vapnik, A training algorithm for optimal margin
  classifiers, in: {Proceedings} of the {5th} {Annual} {Workshop} on
  {Computational} {Learning} {Theory}, 1992, pp. 144--152.

\bibitem{awad2015support}
M.~Awad, R.~Khanna, Support vector regression, in: {Efficient} {Learning}
  {Machines}, Springer, 2015, pp. 67--80.

\bibitem{nguyen2010mining}
H.~V. Nguyen, H.~H. Ang, V.~Gopalkrishnan, Mining outliers with ensemble of
  heterogeneous detectors on random subspaces, in: {Proceedings} of
  {International} {Conference} on {Database} {Systems} {for} {Advanced}
  {Applications}, Springer, 2010, pp. 368--383.

\bibitem{kriegel2011interpreting}
H.~Kriegel, P.~Kroger, E.~Schubert, A.~Zimek, Interpreting and unifying outlier
  scores, in: {Proceedings} of {SIAM} {International} {Conference} on {Data}
  {Mining}, 2011, pp. 13--24.

\bibitem{aggarwal2015theoretical}
C.~C. Aggarwal, S.~Sathe, Theoretical foundations and algorithms for outlier
  ensembles, {ACM} {SIGKDD} {Explorations} {Newsletter} 17~(1) (2015) 24--47.

\bibitem{kaggle}
Kaggle, \href{https://www.kaggle.com/datasets}{Kaggle data repository} (2010).
\newline\urlprefix\url{https://www.kaggle.com/datasets}

\bibitem{campos2016evaluation}
G.~O. Campos, A.~Zimek, J.~Sander, R.~J. Campello, B.~Micenkov{\'a},
  E.~Schubert, I.~Assent, M.~E. Houle, On the evaluation of unsupervised
  outlier detection: measures, datasets, and an empirical study, {Data}
  {Mining} and {Knowledge} {Discovery} 30~(4) (2016) 891--927.

\bibitem{liu2009encyclopedia}
L.~Liu, M.~T. {\"O}zsu, Encyclopedia of database systems, Vol.~6, Springer New
  York, NY, USA, 2009.

\bibitem{FSinR}
A.-R. Francisco, J.-V. Alfonso, A.-A. Antonio, B.~José~Manuel, Fsinr: an
  exhaustive package for feature selection, {Arxiv} {E}-{Prints} (2020)
  arXiv:2002.10330.

\bibitem{ipred}
A.~Peters, T.~Hothorn, ipred: Improved predictors, r package version 0.9-9
  (2019).

\bibitem{glmnet}
J.~Friedman, T.~Hastie, R.~Tibshirani, Regularization paths for generalized
  linear models via coordinate descent, {Journal} of {Statistical} {Software}
  33~(1) (2010) 1--22.

\bibitem{kriegel2008angle}
H.~Kriegel, M.~Schubert, A.~Zimek, Angle-based outlier detection in
  high-dimensional data, in: {Proceedings} of the {14th} {ACM} {SIGKDD}
  {International} {Conference} on {Knowledge} {Discovery} and {Data} {Mining},
  2008, pp. 444--452.

\bibitem{scholkopf2001estimating}
B.~Sch{\"o}lkopf, J.~C. Platt, J.~Shawe-Taylor, A.~J. Smola, R.~C. Williamson,
  Estimating the support of a high-dimensional distribution, {Neural}
  {Computation} 13~(7) (2001) 1443--1471.

\bibitem{dbscan}
M.~Hahsler, M.~Piekenbrock, D.~Doran, {dbscan}: Fast density-based clustering
  with {R}, {Journal} of {Statistical} {Software} 91~(1) (2019) 1--30.

\bibitem{iforest}
F.~T. Liu, IsolationForest: Isolation forest, r package version 0.0-26/r4
  (2009).

\bibitem{abod}
J.~Jimenez, abodOutlier: Angle-based outlier detection, r package version 0.1
  (2015).

\bibitem{e1071}
D.~Meyer, E.~Dimitriadou, K.~Hornik, A.~Weingessel, F.~Leisch, e1071: Misc
  functions of the department of statistics, probability Theory Group
  (Formerly: E1071), TU Wien, r package version 1.7-3 (2019).

\bibitem{HighDimOut}
C.~Fan, HighDimOut: Outlier detection algorithms for high-dimensional data, r
  package version 1.0.0 (2015).

\bibitem{bnlearnpackage}
M.~Scutari, Learning bayesian networks with the {bnlearn} {R} package,
  {Journal} of {Statistical} {Software} 35~(3) (2010) 1--22.

\bibitem{chang2023data}
C.~Chang, J.~Yoon, S.~Arik, M.~Udell, T.~Pfister, Data-efficient and
  interpretable tabular anomaly detection, in: {Proceedings} of the {29th}
  {ACM} {SIGKDD} {Conference} on {Knowledge} {Discovery} and {Data} {Mining},
  2023, pp. 190--201.

\bibitem{han2022adbench}
S.~Han, X.~Hu, H.~Huang, M.~Jiang, Y.~Zhao, Adbench: Anomaly detection
  benchmark, {Advances} in {Neural} {Information} {Processing} {Systems} 35
  (2022) 32142--32159.

\bibitem{bnlearn}
M.~Scutari, \href{http://www.bnlearn.com/bnrepository}{Bayesian network
  repository} (2009).
\newline\urlprefix\url{http://www.bnlearn.com/bnrepository}

\bibitem{lu2018effective}
S.~Lu, L.~Liu, J.~Li, T.~D. Le, Effective outlier detection based on bayesian
  network and proximity, in: {Proceedings} of {2018} {IEEE} {International}
  {Conference} on {Big} {Data} ({Big} {Data}), IEEE, 2018, pp. 134--139.

\bibitem{aggarwal2017outlier}
C.~C. Aggarwal, S.~Sathe, Outlier ensembles: an introduction, Springer, 2017.

\end{thebibliography}

\end{document}